\documentclass[pdflatex,epj]{sn-jnl}% Math and Physical Sciences Numbered Reference Style — REVISED VERSION
\usepackage{graphicx}%
\usepackage{multirow}%
\usepackage{amsmath,amssymb,amsfonts}%
\usepackage{amsthm}%
\usepackage{mathrsfs}%
\usepackage[title]{appendix}%
\usepackage{xcolor}%
\usepackage{textcomp}%
\usepackage{manyfoot}%
\usepackage{algorithm}%
\usepackage{algorithmicx}%
\usepackage{algpseudocode}%
\usepackage{listings}%
\theoremstyle{thmstyleone}%
\theoremstyle{thmstyletwo}%

\theoremstyle{thmstylethree}%

\usepackage[square,numbers,sort&compress]{natbib}%
\usepackage{adjustbox}
\usepackage{tcolorbox}
\makeatletter
\newcommand{\disablepackage}[2]{%
  \disable@package@load{#1}{#2}%
}
\newcommand{\reenablepackage}[1]{%
  \reenable@package@load{#1}%
}
\makeatother

\usepackage{morewrites}
\usepackage{auxhook}  % Better aux file handling for LuaLaTeX
\usepackage{amsfonts}

\usepackage[T1]{fontenc}
\usepackage{graphicx}

\usepackage{mathrsfs} % for \mathscr command
\usepackage{bbm}

\usepackage{lipsum}
\usepackage{xspace}

\usepackage{amsmath}
\usepackage{siunitx}
\usepackage{etoolbox}

\usepackage{bm}
\usepackage{booktabs}\let\cline\cmidrule
\usepackage{multirow}
\usepackage{tabularx}

\usepackage[caption=false]{subfig}% \captionsetup{compatibility=false}
\renewrobustcmd{\bfseries}{\fontseries{b}\selectfont}
\renewrobustcmd{\boldmath}{}
\newrobustcmd{\B}{\bfseries}

\DeclareMathOperator*{\argmin}{arg\,min}

\newcommand{\vtime}{\ensuremath{t}\xspace}
\newcommand{\vround}{\ensuremath{r}\xspace}
\newcommand{\vseller}{\ensuremath{j}\xspace}
\newcommand{\vbuyer}{\ensuremath{i}\xspace}

\newcommand{\vdealprice}{\ensuremath{p_{\vtime, \vseller, \vbuyer}}\xspace}
\newcommand{\vceprice}{\ensuremath{p^*}\xspace}

\newcommand{\vbid}{\ensuremath{b_{i,t}}\xspace}
\newcommand{\vask}{\ensuremath{a_{j,t}}\xspace}

\newcommand{\vbuyervalsymbol}{\ensuremath{\overline{\beta}\xspace}}
\newcommand{\vsellervalsymbol}{\ensuremath{\underline{\sigma}}\xspace}

\newcommand{\vbuyerval}{\ensuremath{\overline{\beta}_{i}}\xspace}
\newcommand{\vsellerval}{\ensuremath{\underline{\sigma}_{j}}\xspace}

\newcommand{\vbuyervalvec}{\ensuremath{\overline{\bm{\beta}}}\xspace}
\newcommand{\vsellervalvec}{\ensuremath{\underline{{\bm\sigma}}}\xspace}

\newcommand{\vquantile}[2]{\ensuremath{q\left(#1, #2\right)}\xspace}
\newcommand{\vquantilevec}[1]{\ensuremath{\mathbf{q}\left(#1\right)}\xspace}
\newcommand{\vgainsoftrade}{g}

\usepackage{cleveref}
\disablepackage{caption}{}
\disablepackage{subcaption}{}

\usepackage[subtle]{savetrees}

\newcommand{\rev}[1]{#1}
\newcommand{\del}[1]{}

\begin{document}

\title[An `Inverse' Experimental Framework to Estimate Market Efficiency]{An `Inverse' Experimental Framework to Estimate Market Efficiency}


\author*[1]{\fnm{Thomas} \sur{Asikis}}\email{thomas@asikis.ai}

\author[1]{\fnm{Heinrich H.} \sur{Nax}}\email{heinrich.nax@uzh.ch}


 \affil*[1]{\orgdiv{Behavioral Game Theory}, \orgname{University of Zurich}, \orgaddress{\street{Andreasstrasse}, \city{Zurich}, \postcode{8050}, \country{Switzerland}}}

% \affil[2]{\orgdiv{Department}, \orgname{Organization}, \orgaddress{\street{Street}, \city{City}, \postcode{10587}, \state{State}, \country{Country}}}

% \affil[3]{\orgdiv{Department}, \orgname{Organization}, \orgaddress{\street{Street}, \city{City}, \postcode{610101}, \state{State}, \country{Country}}}


\abstract{
Digital marketplaces processing billions of dollars annually represent critical infrastructure in sociotechnical ecosystems, yet their performance optimization lacks principled measurement frameworks that can inform algorithmic governance decisions regarding market efficiency and fairness from complex market data.
By looking at orderbook data from double auction markets alone, because bids and asks do not represent true maximum willingnesses to buy and true minimum willingnesses to sell, there is little an economist can say about the market's actual performance in terms of allocative efficiency.
We turn to experimental data to address this issue, `inverting' the standard induced value approach of double auction experiments. 
Our aim is to predict key market features relevant to market efficiency, particularly allocative efficiency, using orderbook data only---specifically bids, asks and price realizations, but not the induced reservation values---as early as possible. Since there is no established model of strategically optimal behavior in these markets, and because orderbook data is highly unstructured, non-stationary and non-linear, we propose quantile-based normalization techniques that help us build general predictive models. We develop and train several models, including linear regressions and gradient boosting trees, leveraging quantile-based input from the underlying supply-demand model. Our models can predict allocative efficiency with reasonable accuracy from the earliest bids and asks, and these predictions improve with additional realized price data. The performance of the prediction techniques varies by target and market type. Our framework holds significant potential for application to real-world market data, offering valuable insights into market efficiency and performance, even prior to any trade realizations.
}

\keywords{competitive \rev{equilibrium} markets, double auctions, market efficiency, machine learning, allocative efficiency}

\maketitle

\section{Motivation}\label{sec:motivation}

As current digital societies increasingly rely on algorithmic systems for economic coordination, 
e.g. from procurement platforms to resource allocation mechanisms, the ability to assess and optimize market efficiency becomes essential for sustainable socio-technical development. 
Traditional economic theory may be difficult to apply when market complexity exceeds analytical tractability, creating a critical gap where traditional approaches can extract meaningful efficiency signals from high-dimensional orderbook data. 
Our framework addresses this challenge by developing machine learning methods that can predict allocative efficiency from observable market patterns, providing the computational tools necessary for evidence-based optimization of digital economic systems in smart societies.

Key determinants of bargaining behavior in markets are often unobservable in real-world data.
In the context of auction, trade and orderbook data, for instance, we are often limited to observing bids and asks, without having access to the traders' corresponding true reservation prices. 
As economists, we are therefore unable to make some of the most crucial evaluations of realized market outcomes in terms of their efficiency, fairness, and welfare properties, which all necessitate knowledge of these unobservables. 
In a bilateral double auction, for example, we could make the most exciting statements from an economists perspective only if, in addition to bids, asks and realized prices, we also knew the true buyer and seller valuations; when trade occurs at a given price we could use such information to assess relative bargaining success and resulting fairness properties of the realized deal, and, conversely, when no trade occurs we would be able to tell whether this was because there were no positive gains of trade between the two parties or whether it was a case of inefficient bargaining and how much the two parties contributed to it relatively.  

\begin{figure}[!ht]
    \centering
    \caption{\emph{The `inverse' approach vis-a-vis the experimental approach.} 
        A conceptual comparison of the classical experimental market approach and our proposed predictive framework. 
        The traditional experimental approach (bottom arrows) follows the established methodology since \citet{smith1962experimental}: 
        experimenter induce private reservation values in controlled laboratory settings, observe the resulting market behavior through bids, asks, and trades, 
        and then evaluate whether these outcomes converge to theoretical Competitive Equilibrium (CE) predictions. 
        This approach has been the cornerstone of experimental economics for validating market theories. 
        Our `inversion' (top arrows) reverses this logic: starting from observable orderbook data alone, 
        we infer the underlying market structure and CE features without access to reservation values, 
        training our models on experimental data where the true values are known for validation.
    }
    {\includegraphics[width=0.9\textwidth]{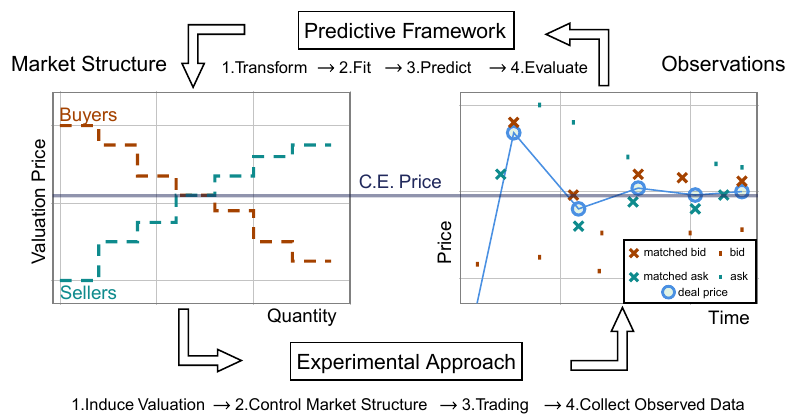}}
    \label{fig:schema}
\end{figure}

In this paper, we propose a novel approach to answer these fundamental efficiency questions in the context of continuous bid-ask double auctions for nondurables a la Vernon \citet{smith1962experimental}. We shall use controlled experimental data from such markets, in particular relying on the data from a large-scale reproduction of \citet{smith1962experimental} by \citet{ikicacompeq}.
We develop a framework that `inverts' the classical analytical approach of experimental studies using standard statistical learning approaches. The classical analytical approach uses the controlled nature of the experimental trade environment to directly test for the convergence of realized market dynamics in terms of Walrasian Competitive Equilibrium (CE). 
We invert this logic as our goal is to predict, based on (initial) orderbook data of realized trade behavior, that is, using data on bids, asks and price realizations---but not reservation prices!---key features of the underlying market structure derived from equilibrium prices and equilibrium gains of trade, and to assess the realized trade behavior in terms of performance vis-a-vis equilibrium. That is, while our models are trained using the controlled nature of experimental data that includes information concerning individual reservation values, our prediction relies on exactly the kind of orderbook data that is typically observable in the field that does not include reservation value information. Our predictive framework is compared with the classical experimental approach schematically in \cref{fig:schema}.

The practical relevance of our framework extends to markets where efficiency assessment remains challenging despite the rich observational data. Consider supply chain procurement platforms where buyers and suppliers engage in repeated negotiations for standardized components. While platform operators observe bid patterns, clearing prices, and allocation outcomes, they cannot directly access suppliers' production costs or buyers' urgency levels, and therefore estimate actual valuation prices. Our inverse approach transforms this observational challenge into an opportunity for inference. By training models on experimental double auction data, where reservation values are known, we develop predictive tools that can assess market efficiency using only observable orderbook information.
This methodology applies to markets with potential assumptions on stable valuations within trading windows and goods that have no resale value in the short term, such as industrial equipment auctions with fixed maintenance schedules, B2B commodity exchanges with contracted delivery dates, on-line ad exchanges or labor platforms matching freelancers to time-bounded projects. In each case, improving allocative efficiency by even 5\% could generate substantial value while ensuring that resources reach their most productive uses. Crucially, our framework acknowledges that real markets deviate from idealized Walrasian assumptions, instead leveraging the specific structure of continuous double auctions to make practical efficiency predictions.

The broader goal of our project is ultimately to propose a method that is able to evaluate the efficiency of a market based on the kind of data that typically would be available, that is, not including information regarding reservation values.\footnote{\rev{This includes bids, asks, realized prices, and market-level metadata such as feedback setting and price rule, but explicitly excludes buyer and seller reservation values.}} Our focus is on markets run as continuous bid-ask double auctions, and in this paper, we use experimental data to develop and train several models. Various approaches out there have pursued that same broader goal in other markets, and other methods have been used before us, which estimate buyers' and sellers' true willingnesses to buy and sell based on real-world data in various ways and contexts. 
To the best of our knowledge, none of them applies directly to markets a la Vernon Smith (1962), i.e., continuous bid-ask double auctions for homogeneous non-durable goods, and none of them has used experimental data to develop their models.
The experimental approach illustrated in the bottom portion of \cref{fig:schema} represents over six decades of methodological development in experimental economics, beginning with \citet{smith1962experimental}. This approach has become the gold standard for testing market theories: researchers induce known reservation values, creating a controlled environment where theoretical predictions can be precisely tested against actual behavior. 
Our inverse approach builds upon this rich experimental tradition, but asks a fundamentally different question: rather than testing whether markets converge to equilibrium given known values, can we infer the underlying value structure from market behavior alone?

Here are examples of interesting papers proposing methods for estimating distributions of valuations from nonexperimental data:
Based on data from English auctions, \citet{haile2003inference} propose one method. 
\citet{larsen2021efficiency} and \citet{larsen2018mechanism} propose related methods for general alternating-offer bargaining settings based on data from markets for used cars. 
Based on data from bilateral sealed-bid k-double auctions (only concerning price realizations), \citet{li2015nonparametric} proposes a nonparametric approach to identify buyer and seller valuations.
See \citet{larsen2021efficiency} for further examples.\footnote{Less related but a great and noteworthy paper in our opinion in this literature is \citet{keniston2011bargaining} analysis of autorickshaw markets with a comparative analysis in terms of efficiency of different price-finding mechanisms (unstructured bargaining versus fixed prices).}
In future work, we would like to run controlled experiments matching these markets to validate, compare and train them. We are not aware of ongoing efforts in this direction.
An important difference between these related efforts and ours is that in the continuous bid-ask double auction, we do not have any theory of rational behavior, which is why we turn to experimental data in combination with machine learning techniques like \citet{fudenberg2019predicting} did successfully to predict early play in matrix games, where such theories were also lacking.

\rev{A separate but related body of work has studied how much of the efficiency observed in double auction markets can be attributed to the market institution itself, as opposed to the rationality of individual traders. \citet{gode1993allocative} showed that zero-intelligence (ZI) traders, who submit random bids subject only to budget constraints, achieve near-maximal allocative efficiency in continuous double auctions, attributing efficiency largely to the institutional structure of the market rather than to strategic behaviour. \citet{cliff1997zero} demonstrated that this finding does not extend to asymmetric supply and demand configurations, where elementary learning rules become necessary for convergence. These results, discussed at length in \citet{friedman2018double}, show how market design parameters such as price rules and information feedback shape efficiency outcomes, and they motivate our choice to include such parameters as features in our predictive models. On the structural-estimation side, \citet{bajari2005structural} used experimental auction data, where true valuations are known, to validate econometric models that recover valuation distributions from observed bids. Our framework shares the strategy of using controlled experimental data as ground truth, but with a different aim: rather than estimating a parametric model of strategic bidding, we train machine learning models to predict equilibrium quantities that depend on unobserved reservation values, taking the perspective of an analyst who observes only orderbook data and market design parameters.}

The kind of direct testing via experimental data that we would like to do for models like \citet{larsen2021efficiency} was pursued by the laboratory approach for testing job search models  (see a review of that approach by \citet{cox1996testing})\footnote{We thank Dan Friedman for pointing us to that literature.}.
Papers in that literature, e.g., \citet{cox1992direct,cox1992tests}, used laboratory experiments to directly test various pre-existent job search models that relied on estimates of unobservables, in particular, estimates of reservation wages. 
The crux of their (and our) experimental approach is that critical features are observable via induced values \citep{Smi76}. In the case of job search, the models and their use of observational data pre-existed their laboratory testing. 
In our setting, we are not aware of any pre-existing models, hence we propose several natural candidates and compare them.
Predicting valuations for data where information concerning valuations is available is not the final step. In future work, we plan to work with observational data.
The novelty of our framework is the `inversion' of the experimental approach. We do not use experiments to test models developed for observational data but rather develop models that are trained to perform well on experimental data and then use them on observational data in a future step.

The remainder of this paper is structured as follows. Next, we outline the underlying market setting.
In section 3, we present the framework and the different models used for prediction. Section 4 contains the results. Section 5 concludes.

\section{Experimental double auctions}

\subsection{Experimental design}\label{sec:experimental:design}
Our study uses data from experimental double auctions by \citet{ikicacompeq}, which is a large-scale reproduction of \citet{smith1962experimental}. The data is well-documented and freely available on public repositories\footnote{\url{https://osf.io/gu62n/} and \url{https://github.com/ikicab/Trading-in-a-Black-Box}} and follows modern standards of computerized experimentation for this kind of market setting, and the data has been used by other papers, for example, in \citet{inoua2022perishable}. 
The dataset summarizes as follows: 

There are in total 104 separate markets, each repeated typically for ten rounds. In each market, there are sets of sellers $S$ and buyers $B$, up to ten of each. Every experimental subject participates either as a seller $\vseller \in S$ or as a buyer $\vbuyer \in B$ with the aim of trading a single item of a homogeneous product once per round $\vround$. Each player is induced with a valuation/reservation price, which determines the on-demand cost of producing the product \vsellerval for a seller and the total available budget \vbuyerval for a buyer. These values are mostly fixed for the whole experiment and private information. Subjects do not know the market size and the distribution of valuations. A seller who trades in a given trading round receives a payoff that is the difference between the price obtained and the reservation price, while a seller who does not trade makes zero profit that round.
Similarly, a buyer who trades makes the difference between the budget and price paid as profit, while a buyer who does not trade forfeits the budget for the round and makes zero profit. Each round lasts for a period of time $T$, during which a continuous double auction is run. During this period, every buyer \vbuyer can submit and overwrite bids $\vbid \leq \vbuyerval$ that are active at time \vtime. Similarly, every seller can submit and overwrite $\vask \geq \vsellerval$ active at time \vtime. Whenever an active bid crosses (that is, is equal to or above) an active ask $\vbid \geq \vask$, then a trade takes place at a realized price  $\vdealprice$ such that $\vbid \geq \vdealprice \geq \vask$ lies between buyer bid \vbuyer and seller ask \vseller at time \vtime. The pair that traded leaves the market until the next round, while the other market participants continue trading until making their own deal or the round ends. For clarity, we use consistent notation throughout: time indices ($\vtime$), trader indices ($\vbuyer$ for buyers, $\vseller$ for sellers), and vector notation (bold) for collections of values.

\begin{figure}
    \centering
    {\caption{\emph{Data summary.} Composition of market experiments and distribution of all 104 markets over the different feedback settings (second ring), price rules (3rd ring), and market sizes (4th ring).
    The numbers in each cell summarize the number of markets ran in this path: e.g., overall, there are 104 markets, of which 29 are `Small' with the `First' price rule and `Full' orderbook access.
    }}
    {\includegraphics[width=0.7\textwidth]{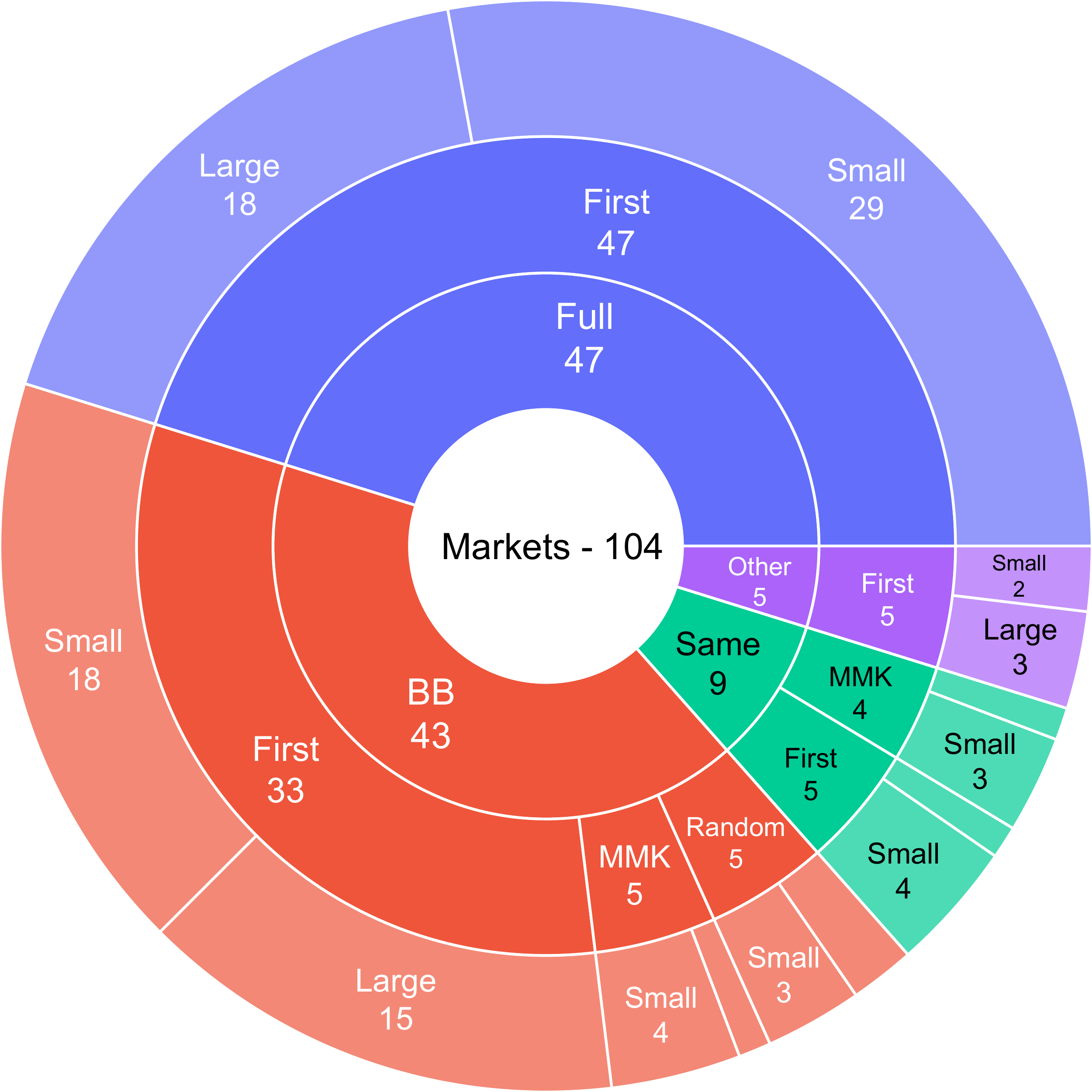}}
    \label{fig:sunburst}
\end{figure}

Different treatments determine the information feedback setting,  the price rule, and the market size. 
Feedback settings (sometimes abbreviated FS) range from `Black Box' (where subjects do not see the bids, asks, and realized prices of others, sometimes abbreviated BB) to `Full' (where subjects have live access to the orderbook and realized prices) with two intermediate treatments `Same' and `Other' (where subjects have access to one side of the orderbook only). There are three price rules (sometimes abbreviated PR), `First' price (where the price is determined by the earlier bid/ask when the two cross), `Random' (where the price is set randomly between bid and ask) and `MMK' (which stands for Matchmaker Keeps, that is, a deal is executed at two prices--the ask of the seller and the bid of the buyer--with the market maker keeping the spread as profit). Some markets are `large' ($\geq$15 traders in total during first round), others `small' ($<$15 traders in total during first round). The most frequent market type is small, with full orderbook feedback and the first price rule. The composition of evaluated markets is illustrated in \cref{fig:sunburst}.

\subsection{Competitive equilibrium and allocative efficiency}\label{sec:competitive:alloc:intro}
The competitive equilibrium (CE) in the kinds of markets described above determines an interval of market-clearing prices such that buyers whose willingness to buy is at or above that price trade with all sellers whose willingness to sell is at or below that price, and every transaction occurs at a unique market-clearing price within that price interval. Crucially, CE outcomes maximize allocative efficiency. 

\emph{Competitive equilibrium price range (CE price range).}  
We denote by  $\left[\underline{\vceprice},\overline{\vceprice}\right]$ the interval of \emph{competitive equilibrium prices}.\footnote{See \citet{Wil85} and \citet{Rus94} for two equivalent functions to find CEPs.} To calculate the CE price range, we denote two vectors of ordered reservation prices, namely the buyer valuations vector $\vbuyervalvec$ with elements in descending order and the ordered seller valuation vector $\vsellervalvec$ in ascending order. We now denote pairs $\vbuyervalsymbol_{k},\vsellervalsymbol_{k}$ of ordered valuations occurring at the same element $k$ in both vectors. The CE prices are found at the earliest occurring index $k^* = \argmin_{k, \vbuyervalsymbol_k -\vsellervalsymbol_k \geq 0}\left(\vbuyervalsymbol_k -\vsellervalsymbol_k\right)$ of both vectors. Considering vector lengths of $l_{\vbuyervalsymbol}, l_{\vsellervalsymbol}$ respectively, we can now calculate the lower bound and upper bounds of CE prices as
\begin{align}
    \label{eq:ce:bounds}
    \underline{\vceprice}   &= \left\{
                                    \begin{array}{ll}
                                          \vsellervalsymbol_{\rev{k^*}}, & l_{\vbuyervalsymbol} = k^*\\
                                           \max \left\{  \vsellervalsymbol_{\rev{k^*}},  \vbuyervalsymbol_{\rev{k^*+1}} \right\}, & \text{otherwise}\\
                                    \end{array}
                                    \right.\\
        \label{eq:ce:upper:bound}
        \overline{\vceprice}   &= \left\{
                                    \begin{array}{ll}
                                          \vbuyervalsymbol_{\rev{k^*}}, & l_{\vsellervalsymbol} = k^*\\
                                           \max \left\{  \vbuyervalsymbol_{\rev{k^*}},  \vsellervalsymbol_{\rev{k^*+1}} \right\}, & \text{otherwise}\\
                                    \end{array}
                                    \right.             
                                    \text{.}
\end{align}

\emph{Gains of Trade (GOT).} 
Realizing all possible trades within the CE price range and in particular at CEP maximizes the total \emph{gains of trade} (GOT) of all pairs $\vbuyer,\vseller \in B \times S$ of buyers and sellers
\begin{align}
    \label{eq:max:optim:gains:trade}
    \vgainsoftrade^{*} &= \max_{B \times S}\sum_{i, j} \left( \vbuyerval - \vsellerval \right), \quad \rev{\{i,j : \vbuyerval \geq \vsellerval \}}\\
                       &= \sum_{k \leq k^*, \vbuyervalsymbol_k -\vsellervalsymbol_k \geq 0} \left(\vbuyervalsymbol_k -\vsellervalsymbol_k\right)\text{.}
\end{align}

\emph{Allocative Efficiency (AE).} 
In a real-world or experimental setting, not all trades will be realized efficiently, i.e., trade may occur outside the equilibrium range, and the realized gains of trade may be inefficient. We can measure the realized gains of trade by summing the differences of reservation prices of every buyer and seller pair  $i,j$, who traded in this round in any arbitrary time $t$
\begin{align}
    \label{eq:real:gains:trade}
    \vgainsoftrade &= \sum_{i, j} \left( \vbuyerval - \vsellerval \right), \quad \{i,j : \vbid \geq \vask \}\text{.}
\end{align}
Following \cref{eq:max:optim:gains:trade,eq:real:gains:trade} we can then measure the realized allocative efficiency (AE) of a trading round up to final time $t=T$ as
\begin{align}
    \label{eq:allocative:efficiency}
    E_\vround &= \frac{\vgainsoftrade}{\vgainsoftrade^{*}}\text{.}
\end{align}
An efficient market achieves maximum AE $E_\vround^* = 1.0$. 
Distance to CE price range or CEP and AE are standard measures to evaluate convergence to CE in experimental double auctions dating back to \citet{smith1962experimental} (see also \citet{ikicacompeq}). 
We note that allocative efficiency is calculated via the non-linear operators of \cref{eq:allocative:efficiency,eq:real:gains:trade,eq:ce:bounds} over reservation prices. Assuming observable bids and asks converge to the true reservation prices over time, we motivate the use of non-linear predictive models over linear ones for the estimation of AE. On the contrary, linear models seem to perform much better when predicting CEP, as shown in related work~\citet{ikicacompeq} and we present in later in \cref{appdx:section:cep}.

\section{Methods}

In this section, we showcase the design of a predictive framework that uses both traditional methods and machine learning, which we called the inverse approach.
Recall, based on orderbook data, our goal is to predict as soon as possible the underlying market's AE and CEP that the market will achieve.
By contrast, the standard experimental approach was to induce values and then to test whether realized market outcomes conform with CE predictions. We will evaluate two classes of methods to make such predictions based on linear and non-linear regressions.

\subsection{Relevant Data and Design Challenges}\label{sec:dataset}

\subsubsection{Observables}
Our goal is to infer the underlying market structure using a model that relies only on the kind of choice data available for a market analyst and using the information regarding induced values available to the experimenter only as learning labels for training and validation of the model. Note that the experimenter knows more than the analyst, who might know more than the subject/trader (e.g.,the experimenter sees the whole history of the orderbook, which a subject might only see when actively trading in the market). 
To further elaborate, those three perspectives have different information access:
\begin{itemize}
\item The \emph{experimenter} observes everything: complete orderbook history, all induced values, and experimental parameters. In real markets, this omniscient view does not exist, which is precisely why we need our predictive framework.
\item The \emph{analyst} (our model's perspective) sees only market data: complete orderbook history, realized prices, and market rules, but crucially \emph{not} the reservation values. This mirrors real-world market observers such as:
\begin{itemize}
    \item A platform operator (e.g., an ad exchange monitoring bid-ask spreads for advertising slots).
    \item A regulatory authority assessing market efficiency.
    \item An academic researcher studying market structure.
    \item A market surveillance system detecting anomalies.
\end{itemize}
\item The \emph{subject/trader} has the most limited view: often only the current orderbook state (depending on feedback setting), their own reservation value, and their own trading history. In real markets, these would be:
\begin{itemize}
    \item Individual advertisers bidding for ad slots (knowing only their own campaign ROI).
    \item Sellers on a marketplace platform (knowing only their own costs).
    \item Traders in commodity markets (knowing only their own inventory needs).
\end{itemize}
\end{itemize}
At a time of prediction $\tau$, we therefore consider that only the following data are available as inputs to our models (taking the analyst's perspective):
\\
 
\emph{Observables:}
\begin{itemize}
     \item Bid-ask data: All bids and asks that have been submitted: $\vbid,\vask$ with $t \leq \tau$
     \item Realized prices: From the order book, we also know in particular every realized deal price, $\vdealprice, t \leq \tau$, and the count of realized deals, $n_\tau = \sum \mathbbm{1}_{t \leq \tau}(\vdealprice)$.
     \item Timing: The current time $\tau$ since the start of the round and the round number \vround are also considered known.
     \item Feedback/information available: The feedback setting and the price rule are also known and can be modeled as categorical variables.
\end{itemize}

\rev{Market design parameters such as the feedback setting and price rule are publicly known rules of the market mechanism and are therefore observable to the analyst, as they would be to a platform operator or regulator.}

By contrast, the following data are considered unobserved as inputs for predictive modeling:
\\

\emph{Unobservables:}
\begin{itemize}
     \item Valuations: Reservation prices and any statistic relevant to reservation prices, for example, the relevant price scale.
     \item Market size: Total number of participants and total number of assigned sellers/buyers.
     \item Activity: The participants joined later or dropped early in the game.
     \item Market structure: Any information related to market structure, e.g., which side has more participants than the other.
\end{itemize}
Given that the sets of bids and asks change dynamically over time and the number of total participants is unknown and potentially not fixed across games, it would be challenging to use such sets in standard learning algorithms, which often rely on structured inputs with predetermined dimensions, such as vectors, matrices, or tensors.
Furthermore, past research on machine learning in finance~\citep{wu2021robust,briola2020deep,nousi2019machine} has shown that direct use of orderbook information is not easily supported by standard machine learning algorithms as the price levels may change in time and the range of price levels is unknown.
These papers also show that the relationship between volume and price can be distorted by neural network activations, even in more complex machine learning models.

\rev{More broadly, the machine learning literature on limit order books has developed progressively richer representations for financial market data. \citet{kercheval2015modelling} proposed a feature set based on fixed-depth price and volume snapshots at multiple levels of the order book to predict mid-price movements using support vector machines. \citet{ntakaris2018benchmark} introduced a benchmark dataset for mid-price forecasting with normalised LOB representations and systematic baseline comparisons. \citet{zhang2019deeplob} extended these approaches with deep convolutional neural networks operating on multi-level LOB snapshots and reported the best benchmark performance at the time of publication. These methods are designed for financial limit order books with continuous trading of durable assets, deep liquidity across many price levels, and no concept of trading rounds or equilibrium. The experimental double auction markets we study differ in fundamental ways: each trader holds a single unit of a homogeneous non-durable good, trades at most once per round, and the market restarts with fresh rounds under the same induced valuations. Fixed-depth LOB representations are therefore not directly applicable, as the number of active orders is small and varies across markets. Our quantile-based representation accommodates this variability by summarising the distribution of bids and asks at each prediction time regardless of market size, while remaining agnostic to the absolute price scale.}

\begin{figure}[htbp!]
    \centering
    \includegraphics[width=0.8\linewidth]{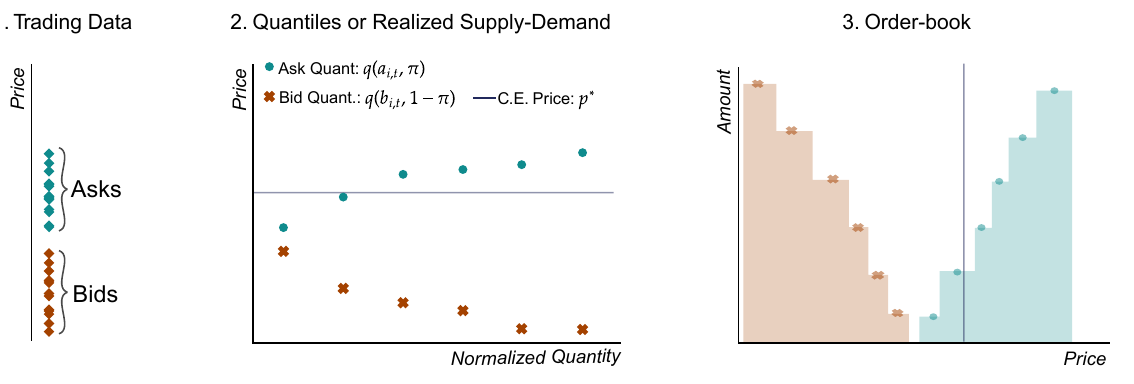}
    \caption{\emph{From orderbook data to quantiles.} The figure illustrates the connection of quantiles of observed bids and asks up to prediction time with supply-demand curves and orderbook status.
             Note that the deciles resemble the normalized quantities of demand and supply, where instead of using the cumulative number of players on the horizontal axis (see left of \cref{fig:schema}, we instead use the cumulative sum of players divided by the total number of players.)
             }
    \label{fig:quant}
\end{figure}

\subsubsection{Quantile Representation of Supply-Demand}
In this article, building on the results of the papers as mentioned earlier, we take a slightly different approach to represent information coming from player actions in a more structured manner: instead of using raw orderbook bids and asks, we transform the data into quantiles $\vquantile{\vbid}{\pi}, \vquantile{\vask}{\pi}$ on the sets of active bids and asks at probability $\pi$.
In particular, we choose $10$ equally spaced deciles (denoted by D1-D10) and create a decile vector of bids (where D5 includes the median), including also the minimum values in the beginning, which we abbreviate as 0\textsuperscript{th} decile (D0):
\begin{equation}
    \vquantilevec{\vbid} = \begin{bmatrix}  \vquantile{\vbid}{0.0} & \vquantile{\vbid}{0.1} & \ldots & \vquantile{\vbid}{1.0}\end{bmatrix}
\end{equation}
Similarly, we can construct the decile vector of sellers' asks $\vquantilevec{\vask}$.
Treating the probabilities $\pi$ as normalized quantities, for buyers, the flipped-order decile vector
\begin{equation*}
    \text{flip}\left(\vquantilevec{\vbid}\right) = \begin{bmatrix}  \vquantile{\vbid}{1.0} & \vquantile{\vbid}{0.9} & \ldots & \vquantile{\vbid}{0.0}\end{bmatrix} 
\end{equation*}
expresses the realized demand, while, for sellers, the decile vector represents the realized supply.
Finally, the combination of these vectors is a proxy of the orderbook information, as illustrated in the connection between quantiles, supply and demand, and order book in Figure \ref{fig:quant}.

\subsubsection{Scale and Normalization}
The next challenge is the lack of information about reservation prices, which prevents us from determining the proximity of a bid or ask to its respective reservation price.
Additionally, to generalize the predictive models to unobserved experimental or real-world data, we cannot rely on knowing the trading price scale, especially before any deals, bids, or asks occur.
For example, even when considering experimental data, the current dataset uses experimental data on the scale of hundreds. In contrast, we plan to use in future work another dataset with different scales, such as \citet{lin2020general}.

Normalization can be achieved in two manners, either we use models that calculated scale-invariant coefficients, e.g. linear models with price features and prediction targets (e.g. CEP), as well as no-intercept terms to capture scale, or we can normalize the price related inputs, such as bid and ask quantiles using an interquartile range (IQR) normalization is calculated by taking the median and interquartile range (35th and 65th quantiles) values of the concatenated vector of bids and asks  $X_{q} = \begin{smallmatrix} \vquantilevec{\vbid} & \vquantilevec{\vask}\end{smallmatrix}$ and calculating
\begin{align}
\label{eq:quantile:norm}
   X_{q}^{\dagger} &= \dfrac{X_{q} - \text{median}\left(X_{q}\right)}{\text{iqr}_{\left[0.35, 0.65\right]}\left(X_{q}\right)}\text{.}
\end{align} 

For a predictive model to generalize to such new data, we would need to remove the effects of pricing scale to its prediction, meaning that any input feature related to price should be first normalized or there should be no model coefficients for input features that are in a different scale than the prediction targets. When predicting allocative efficiency, the output range is always in $\left[ 0,1 \right]$, so in this case, all relevant price input features need to be normalized first.

\subsubsection{Model Fitting and Evaluation}\label{sec:model:fitting}
Perform the experimental process, as shown in \cref{fig:schema}, and then collect realized market data and extract order book information.
Next, we fit an estimator on 50\% of market data for “training” a model. 
The data are split so that for each treatment, half of the games belong to the training set, and the other half belong to the test set.
The learning objective minimizes a loss w.r.t. competitive equilibrium price or allocative efficiency.
To perform hyper-parameter selection and model design, we use a random 80\% of the training set for fitting and the remaining 20\% to select the best-performing parametrization (validation set).
Next, we evaluate the estimators' performance and present the results in the remaining test set.

\emph{Main prediction assessment criterion: Median Absolute Percentage Error (Median-APE/ MAPE)).} \\ 
For evaluation between a prediction $\hat{y}\in\mathbb{R}$ and a target value $y\in\mathbb{R}$. The target value is the true quantity of a property that we may only be able to measure or calculate only during an experimental process or after deals have been completed. For evaluation we use the median absolute precision error between predicted and target values as:
\begin{align}\label{eq:median:ape}
    \text{Median-APE} = \text{med}\left|\frac{y-\hat{y}}{y}\right|
\end{align}
If the target value is zero (as may be the case in allocative efficiency), we replace the denominator with the predicted value if it is non-zero or one otherwise.
For example in the presented results, based on evaluation metric we use $\hat{y} = E$ or $\hat{y} = \vceprice$.
To select hyper-parameter grids and perform feature selection, we use small validation sets for training from the training set of 4-5 random training-test splits.
The main challenge with the current evaluation procedure is that the entirety of the dataset has been used in past studies~\citep{ikicacompeq} that we draw knowledge to select the hyper-parameters of the proposed predictive models.
To alleviate the effects of designer bias in our results, we use randomly sampled $50$ splits between the training and test sets, where we refit each model (including any hyper-parameter selection process) on 50\% of the dataset and test on the remaining 50\%.
In the future, we plan to use data from other studies (e.g., \citet{lin2020general}) on the fitted models on the whole dataset to determine better estimates for the out-of-sample performance of the proposed methods.
Regarding the model fitting procedure, we choose between two possible routines. Either we fit the model on the whole training set, especially for highly complex non-linear models that can utilize categorical and time-related features (e.g., the round number), or we fit the model on the training data up to a given training round or deal price for different categories (grouped regressions) and then test on the corresponding round data in the test set.
The latter procedure is often preferred for simpler models, such as linear regression.

Our general \rev{prediction} framework can be \rev{summarized} in the following manner:
\begin{tcolorbox}[colback=gray!5!white,colframe=gray!75!black,title=Summary: Efficiency Prediction Framework]
    \textbf{Input:} Orderbook data (bids $b_{i,t}$, asks $a_{j,t}$, prices $p_{t,j,i}$) \\
    \textbf{Transform:} Convert to quantile vectors $\mathbf{q}(b), \mathbf{q}(a)$ \\
    \textbf{Normalize:} Apply IQR normalization (Eq. \ref{eq:quantile:norm}) - Make inputs and outputs scale invariant when required.\\
    \textbf{Predict:} $\hat{E}$ (efficiency) or $\hat{p}$ (price) using Linear/GBT models \\
    \textbf{Output:} Market efficiency assessment without knowing reservation values
\end{tcolorbox}

\subsection{Efficiency prediction of AE}

\subsubsection{Efficient Market Hypothesis (EMH)}
In this case the efficient market hypothesis estimates an allocative efficiency of $1$ for all samples, as we assume that all buyers and sellers with optimal gains of trade can trade in an efficient market.

\subsubsection{`Corrected' EMH (CEMH)}\label{sec:cemh:ae}
There can be experimental markets ---also present in our data--- that contain very few to no deals in the first round. 
Using the EMH in such cases may lead to very inaccurate predictions of market efficiency.
Since the model targets have the same scale, regardless of the experimental setting, we can now construct a non-parametric estimator, relying on allocative statistics in our training set.
When the training set is large enough, we can use the median allocative efficiency across different training subsets partitioned by categorical and time-related features as a predictor.
This non-parametric model calculates the median AE of each training sample group by the price rule, feedback setting, and number of observed deals per round.
Let's note here that the median operator may not have an equivalent sum-product form, so it cannot be separated into a product of two vectors for any vector length larger than 2.
Furthermore, the partitioning operator may also become non-linear, so it is not clear whether the CEMH model for predicting AE is a linear model.
As a result, we end up with a "new" non-linear model for Corrected Efficient Market Hypothesis that aims to capture statistical macro-patterns in the training data in the hope that they persist in the test data.
The main disadvantages of this model is that (i) it cannot extrapolate estimations for test samples with non-observed partitioning variable values, e.g., when a previously unobserved feedback setting is used, (ii) it doesn't use available price data. 

\subsubsection{Robust Linear Regression on the Orderbook (OB-RLM)}\label{sec:obrlm:ae}
The next model we use can be formalized as a linear model over normalized bids/asks quantiles and deal prices and to observed realized prices $n$ up to prediction time:
\begin{equation}
    \label{eq:ob:rlm:loss}
    \hat{E} = \sum_{\pi} \alpha_{a, \vround, \pi} (\vquantile{\vask}{\pi})^{\dagger} + \sum_{\pi} \alpha_{b, \vround, \pi} (\vquantile{\vbid}{\pi})^{\dagger} 
    + \gamma_{\vround} \vdealprice^{\dagger} + \delta_{\vround} n + \xi
    \text{.}
\end{equation}
Since this model uses normalized input features and the outputs are invariant to the price scale, we now include the intercept term $\xi$, which also has a greater impact on the accuracy of the prediction.
We now note that our aim is to learn a vector of linear regression coefficients:
\begin{equation*}
    \bm{{\alpha}}_{\vround} = \begin{bmatrix} \alpha_{a, \vround, 0.0} & \alpha_{a, \vround, 0.1} & \ldots & \alpha_{b, \vround, 1.0},
     \gamma_{\vround}, \delta_{\vround}, \xi
    \end{bmatrix}
\end{equation*}

As human players are not always expected to play rationally, we observe that some bid and ask prices might have values that lie further away from their expected values, and the fitted regression coefficients may be affected by such values.
For example, if a seller tries to experiment with their first action, they can submit an ask of two orders of magnitude larger than their reservation price.
In such a case, the current model would need to assign very low coefficient values only for that input sample to avoid a substantial increase in the average loss calculation.
We denote this model as the Orderbook Robust Linear Model (OB-RLM).

We also consider scale effects of input prices to allocative efficiency, i.e., the scale of predicted allocative efficiency should be independent of the input price scales.
Therefore, we normalize the input bids and asks using the same interquartile range normalization discussed in \cref{sec:gbt:price}.
For AE prediction, given that the outputs are always between $E\in[0,1]$, we clip the model outputs respectively to avoid estimating invalid values of allocative efficiency.
After initial testing, we append the realized price $\vdealprice$ to the quantile input vector $X_q$ and calculate the appropriate normalization coefficients.
We also now include a grouping variable for the feedback setting and the unique deals per round and fit a model for each round. \rev{Concretely, the fit is partitioned by feedback setting within each round (consistent with \cref{tab:model:features}), with the bid and ask decile vectors, $\vdealprice$ and $n$ entering as direct inputs and an intercept term. The round dependence is obtained by restricting the training data to rounds up to the current one.}

\subsubsection{Gradient Boosting Trees (GBT)}\label{sec:gbt:ae}
Gradient Boosting Trees (GBT) are successful for non-linear regression tasks due to their iterative nature, ability to handle complex relationships, and robustness to noise. 
GBTs are an ensemble method that builds on weak learners~\citep{Ferreira2012} (typically decision trees) by sequentially fitting new models to correct the error residuals of the previous ones.
We note that GBT also partitions the data, similar to the EMH approach.
This iterative error correction effectively allows the model to focus on areas of the dataset where it is currently performing poorly. 
This process is guided by minimizing a differentiable loss function, achieved through gradient descent optimization~\citep{friedman2001greedy}. 
The combined strength of the ensemble of weak learners helps to balance bias and variance, leading to improved generalization performance~\citep{ganjisaffar2011bagging}.
GBT is relatively robust to outliers and can handle missing values without explicit imputation, thanks to the ensemble approach. 
Despite these strengths, GBTs can be computationally intensive, particularly with larger datasets or deeper trees, and may suffer from overfitting if not properly tuned. 
Additionally, the complexity of the ensemble structure can make the model less interpretable, which, in our case, might reduce the understanding of underlying market dynamics and their connection to the predicted market efficiency measures. Since we expect a non-linear relationship between allocative efficiency and observed input features, we choose a hyper-parameter grid with deeper and wider trees, effectively yielding more complex models.

\subsubsection{Model Comparison Overview}
An overview of the proposed models from above can be found in \cref{tab:model:comparison}. The same models are used for CEP \rev{prediction} in the following section:
\begin{table}[ht]
    \centering
    \caption{Comparison of predictive models for efficiency estimation, with our analysis on model characteristics.}
    \label{tab:model:comparison}
    \footnotesize
    \setlength\tabcolsep{4pt}
    \begin{tabular}{lcccp{4.0cm}}
        \toprule
        \textbf{Model} & \textbf{Expressiveness} & \textbf{Interpretability} & \textbf{Use Case} & \textbf{Key Advantage} \\
        \midrule
        EMH & Low & High & Baseline & Simple benchmark \\
        CEMH & Low & High & Grouped & Captures equilibrium phase \\
        OB-RLM & Medium & High & With prices & Interpretable orderbook use \\
        GBT & High & Medium & Early pred. & Non-linear, transient phase \\
        \bottomrule
    \end{tabular}
\end{table}
Based on \cref{tab:model:comparison}, the model choice reflects a fundamental trade-off: linear models (EMH, CEMH, OB-RLM) offer clear interpretability through coefficients that directly show how bid/ask quantiles influence predictions, while the non-linear GBT model better captures the complex dynamics of early market formation at the cost of interpretability. As shown in our results, this trade-off manifests differently for price versus efficiency prediction.

\rev{The inputs entering each fitted model are summarised in \cref{tab:model:features}. We distinguish three groups: orderbook features (the running bid and ask quantiles $\vquantile{\vbid}{\pi}$ and $\vquantile{\vask}{\pi}$), the last realised price $\vdealprice$, and experimental-protocol features (the feedback setting, the price rule, the round index $\vround$ and the number of deals observed so far in the round $n$). Protocol features enter EMH, CEMH and OB-RLM as grouping or partitioning variables (separate fits per partition) and GBT as direct inputs. The ablations in \cref{sec:appendix:ablations} remove the protocol or the realised-price inputs and refit against this reference.}
\begin{table}[ht]
    \centering
    \caption{\rev{Input features used by each fitted model, separately for allocative efficiency (AE) and competitive equilibrium price (CEP). Orderbook = running bid/ask quantiles $\vquantile{\vbid}{\pi}, \vquantile{\vask}{\pi}$. Deal price = last realised deal price $\vdealprice$. Feedback and Price rule are categorical treatment descriptors. $\vround$ is the round index and $n$ is the number of deals observed in the current round. A symbol in parentheses indicates that the variable is used as a partitioning/grouping variable rather than a direct linear input.}}
    \label{tab:model:features}
    \scriptsize
    \setlength\tabcolsep{2pt}
    \begin{tabular}{l l ccccccc}
        \toprule
        Target & Model & Orderbook & Deal price & Feedback & Price rule & $\vround$ & $n$ & Intercept \\
        \midrule
        \multirow{4}{*}{AE}
        & EMH    &   &   &   &   &   &   & trivially $1.0$ \\
        & CEMH   &   &   & ($\bullet$) & ($\bullet$) & ($\bullet$) & ($\bullet$) &   \\
        & OB-RLM & $\bullet$ & $\bullet$ & ($\bullet$) &   & (partition) & $\bullet$ & $\bullet$ \\
        & GBT    & $\bullet$ &   & $\bullet$ & $\bullet$ & $\bullet$ & $\bullet$ &   \\
        \midrule
        \multirow{4}{*}{CEP}
        & EMH    &   & $\bullet$ &   &   &   &   &   \\
        & CEMH   &   & $\bullet$ & ($\bullet$) & ($\bullet$) &   & (partition) &   \\
        & OB-RLM & $\bullet$ &   &   &   & (partition) &   &   \\
        & GBT    & $\bullet$ &   & $\bullet$ & $\bullet$ & $\bullet$ & $\bullet$ &   \\
        \bottomrule
    \end{tabular}
\end{table}

\subsection{Price prediction of CEP}\label{appdx:section:cep}

\emph{Competitive equilibrium midprice (CEP).} In this article, we focus on the midpoint of the CE price range, which we refer to as CEP and denote by $\vceprice = \frac{\underline{\vceprice} + \overline{\vceprice}}{2}$. This choice is motivated by two main reasons, namely (i) the tight CE price range, which is often within $2\%$ of the relative CEP, and (ii) the ability of streamlined prediction models to achieve better performance at point estimation tasks compared to interval prediction. 

We run and compare the following four models for predicting the CEP:
\begin{itemize}
    \item \emph{Assuming market efficiency:} a non-parametric model that uses the last observed realized price to estimate the competitive equilibrium price based on the efficient market hypothesis (EMH).
    \item \emph{Adjusted price estimate:} a parametric correction of the EMH model, where we fit a linear estimator on the last observed price.
    \item \emph{Robust regression:} a linear robust estimator that we fit on the decile vectors.
    \item \emph{Gradient boosting tree:} a quantile regression gradient boosting tree.
\end{itemize}   

We shall now make these four approaches explicit for the goal of CEP prediction. 
When our prediction target is the competitive equilibrium price, we need to either normalize both input features and targets or that all coefficients are multiplied with price input features that are expected to be in a similar range with reservation prices.

\subsubsection{Efficient Market Hypothesis (EMH)}
With some abuse of terminology of what is usually meant with the efficiency market hypothesis (EMH) \citep{fama1970efficient}, this model assumes that the realized prices reflect the reality of competitive equilibrium.
This non-parametric estimator will then be
\begin{equation}
    \hat{p} = \vdealprice \text{.}
\end{equation}
Here, we observe that this model is scale-independent, as the latest observed price is expected to be close to competitive equilibrium price across all possible experiments. 
    
\subsubsection{`Corrected' EMH (CEMH)}\label{sec:cemh:price}
Given the results on price level dynamics from \citet{ikicacompeq}, we know that the EMH model has lower performance at the beginning of the equilibration process, especially when only a few deal prices are available in the initial trading rounds.
Hence, accounting for the convergence trend that is identified in \citet{ikicacompeq} (and in earlier studies dating back to \citet{smith1962experimental}), which has a bias favoring buyers initially, our second model allows for a time-dependent correction term to improve performance: realized prices are scaled by a coefficient estimated separately per round $\vround$ and \rev{per number of observed deal prices} $n$\rev{,} resulting in the following form:
\begin{equation}
    \hat{p} = \alpha_{n, \vround} \vdealprice
\end{equation}
\rev{In our implementation the scaling coefficient is partitioned by the treatment descriptors and by $n$, consistent with \cref{tab:model:features}.}

This new `corrected' EMH (CEMH) model strongly resembles the linear regression model~\citep{james2013introduction}, but we observe that the intercept term is omitted.
The intercept term captures constant effects caused by fixed scale offsets in the data, potentially capturing the scale of the underlying experimental data used for fitting.
In our case, capturing such effects would decrease model performance in unobserved markets, as the scale of the prices may vastly vary from market to market, e.g., markets with different exchange currencies. 
Therefore, discarding any coefficients of terms with constant effects in our proposed models is essential.
Finally, we fit the model using a robust objective \cref{eq:huber:loss} (Huber Loss, \citealt{huber1964robust} ) for a single sample to remedy the possible problem of ordinary least squares coefficients being affected by outlier values in training samples.

\subsubsection{Robust Linear Regression on the Orderbook (OB-RLM)}\label{sec:obrlm:price}
The previous models rely on a realized price to make a prediction.
However, when using the available bid and ask values there is an opportunity to predict the competitive equilibrium price before any price realization. 
Therefore, the next model we use can be formalized as
\begin{equation}
    \hat{p} = \sum_{\pi} \alpha_{a, n, \vround, \pi} \vquantile{\vask}{\pi} + \sum_{\pi} \alpha_{b, n, \vround, \pi} \vquantile{\vbid}{\pi} \text{.}
\end{equation}
We omit the intercept term in this model to avoid learning coefficients dependent on price scaling.
We now note that our aim is to learn a vector of linear regression coefficients:
\begin{equation*}
    \bm{{\alpha}}_{\vround, n} = \begin{bmatrix} \alpha_{a, n, \vround, 0.0} & \alpha_{a, n, \vround, 0.1} & \ldots & \alpha_{b, n, \vround, 1.0}  \end{bmatrix}
\end{equation*}
\rev{The fitted model regresses $\vceprice$ on the bid and ask decile vectors only, without an intercept; no realised deal price, treatment descriptor, $n$ or $\vround$ enters as a direct feature. The round index is used only to partition the training data, so that rounds beyond the current one do not contribute to fitting.}
In this model, the effect of outlier values is even more considerable when predicting CEMH. Hence, we also fit the model using a robust loss objective~\citet{james2013introduction} as shown in \cref{eq:huber:loss}, when predicting normalized prices $p$:
\begin{equation}
\label{eq:huber:loss}
    J(\vceprice, \hat{p}; \bm{{\alpha}}_{\vround, n}) = 
    \left\{
    \begin{array}{ll}
        \left(\vceprice - \hat{p}\right)^2, & |\vceprice - \hat{p}| \leq \rev{1.345} \\
        |\vceprice - \hat{p}|, & \text{otherwise}\\
    \end{array}
    \right. \text{.}
\end{equation}
We note that the threshold value of $\rev{1.345}$ is the default in the \texttt{statsmodels} implementation of this loss that we use in our experiments.
For multiple samples, we can take the average overall calculated loss values.

In the case of allocative efficiency, the robust loss becomes the standard ordinary least squares loss, as allocative efficiency is bounded in $[0, 1]$.
As human players are not always expected to play rationally, we observe that some bid and ask prices might have values that lie further away from their expected values, and the fitted regression coefficients may be affected by such values.
For example, if a seller tries to experiment with their first action, they can submit an ask of two orders of magnitude larger than their reservation price.
In such a case, the current model would need to assign very low coefficient values only for that input sample to avoid a substantial increase in the average loss calculation.
We denote this model as the Orderbook robust Linear Model (OB-RLM).

\subsubsection{Gradient Boosting Trees (GBT)}\label{sec:gbt:price}

We use GBT to estimate the normalized competitive equilibrium price based on the following inputs: the normalized bids/asks, the feedback setting, and price-rule categorical variables, as well as the current round and the number of observed deal prices in this round. 
The model fits once per train-test split on the whole training set.
We also scale the target CEP values to the normalized range by using the previously calculated values
\begin{align}
\label{eq:quantile:norm:target}
   p^{\dagger} &= \dfrac{\vceprice - \text{median}\left(X_{q}\right)}{\text{iqr}_{\left[0.35, 0.65\right]}\left(X_{q}\right)}\text{.}
\end{align} 
Finally, we use the median quantile loss to predict the normalized targets $p^{\dagger}$, which we consider as a better alternative for robust loss in GBTs, namely:
\begin{align}
\label{eq:quantile:loss}
    J(p^{\dagger}, \hat{p^{\dagger}}) &= \sum_n (c - \mathbbm{1}(p^{\dagger}_n \leq \hat{p^{\dagger}}_n))(p^{\dagger}_n - \hat{p^{\dagger}}_n)\text{,}
\end{align} 
where $c=0.5$ is a user-defined coefficient, denoting the quantile to use for the loss calculation.
During inference time, we can retrieve the non-normalized price prediction via the inverse transform:
\begin{align}
\label{eq:quantile:norm:target:inverse}
  \hat{\vceprice} &= \hat{p^{\dagger}}\cdot\text{iqr}_{\left[0.35, 0.65\right]}\left(X_{q}\right) + \text{median}\left(X_{q}\right)\text{.}
\end{align}

\section{Results}
In this section, we compare models in terms of test set performance. 
We take the original dataset discussed in \cref{sec:dataset} and calculate the Median APE/MAPE (\cref{{eq:median:ape}}) in terms of CEP and AE across all test samples. 

\subsection{AE Prediction Performance and Analysis}\label{sec:results:analysis}

\begin{table}[htb!]
    \caption{\emph{Left: Allocative efficiency prediction error before and after initial price realizations.} AE test set performance for per-round AE values calculated separately prior to any price realization and once at least one price is realized. Bold is better. 
    \emph{Right: Allocative efficiency prediction error for small and large markets.} MAPE test set performance for per-round AE values calculated separately for `Small' and `Large' markets. Bold is better.
    This table doesn't provide confidence intervals as we performed paired-non parametric tests to find statistically significant out-performance between models. The results of these tests are found in \cref{tab:ae_ape:wilcoxon}.
    }
    \centering
    \hrule
    \footnotesize
    \bgroup
    \def\arraystretch{1.0}
    \setlength\tabcolsep{3pt}
    \begin{tabular}{c || c | c c c c | c | c c c c}
        & \multicolumn{5}{c|}{\textbf{Price Realizations Analysis (Left)}} & \multicolumn{5}{c}{\textbf{Market Size Analysis (Right)}} \\
        Round & \# realizations & EMH & CEMH & OB-RLM & GBT & Market size & EMH & CEMH & OB-RLM & GBT \\
        \midrule
        \multirow[c]{2}{*}{1} & 0 & 0.500 & 0.173 & 0.216 & \B 0.168 & Small & 0.231 & 0.181 & 0.185 & \B 0.131 \\
        & 1+ & 0.390 & 0.116 & 0.164 & \B 0.104 & Large & 0.394 & 0.105 & 0.166 & \B 0.102 \\
        \cline{1-6} \cline{7-11}
        \multirow[c]{2}{*}{2+} & 0 & 0.171 & \B 0.072 & 0.120 & 0.077 & Small & 0.118 & \B 0.060 & 0.107 & 0.067 \\
        & 1+ & 0.134 & 0.068 & 0.093 & \B 0.066 & Large & 0.148 & 0.070 & 0.092 & \B 0.067 \\
    \end{tabular}
    \egroup
    \label{tab:ae_ape}
    \hrule
\end{table}

\begin{figure}[htb!]
    \centering
    \subfloat[Prior to any price realization]{
        \includegraphics[width=0.45\linewidth]{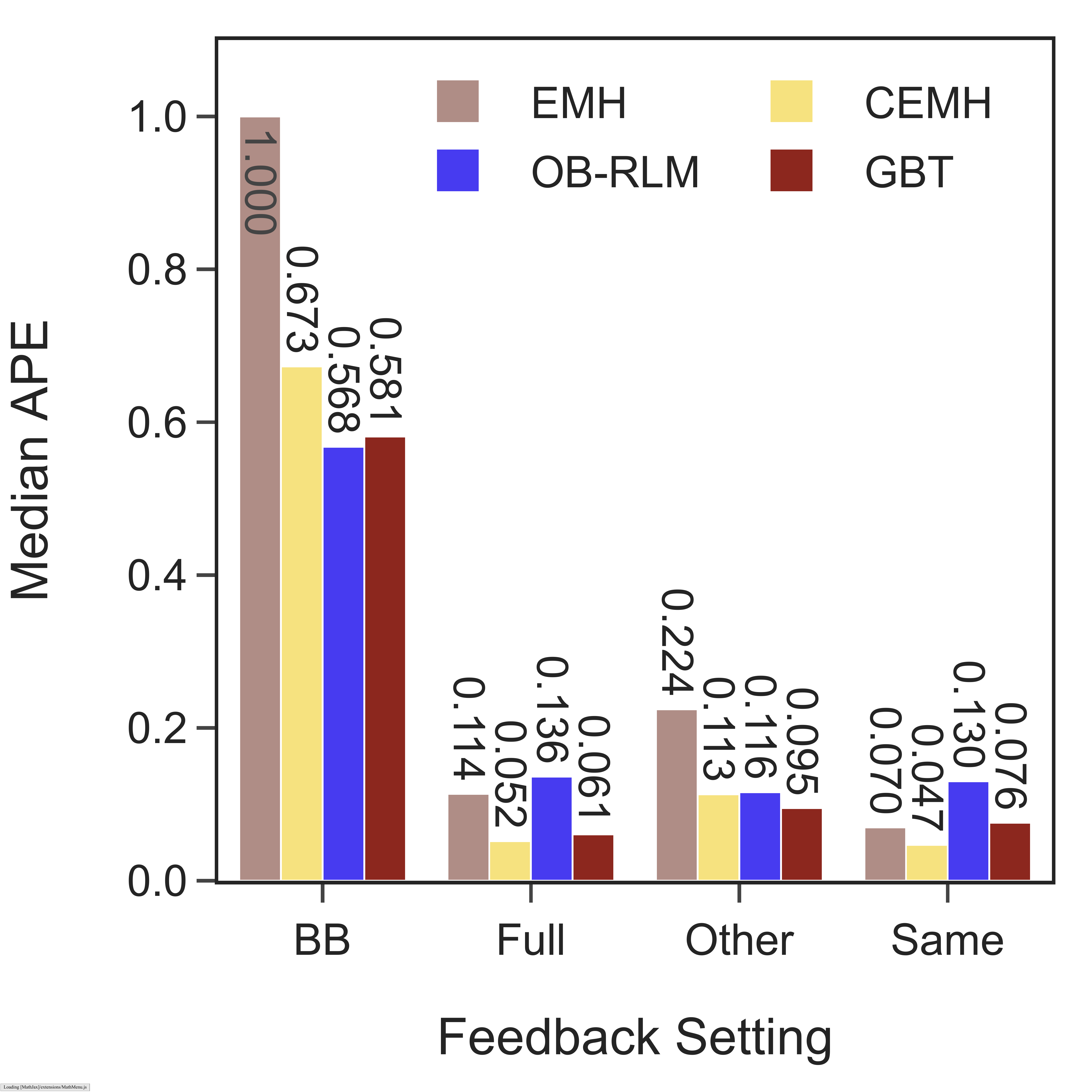}
        \label{fig:ae:setting:ape:0}
    }
    \hfill
    \subfloat[At least one price realization]{
        \includegraphics[width=0.45\linewidth]{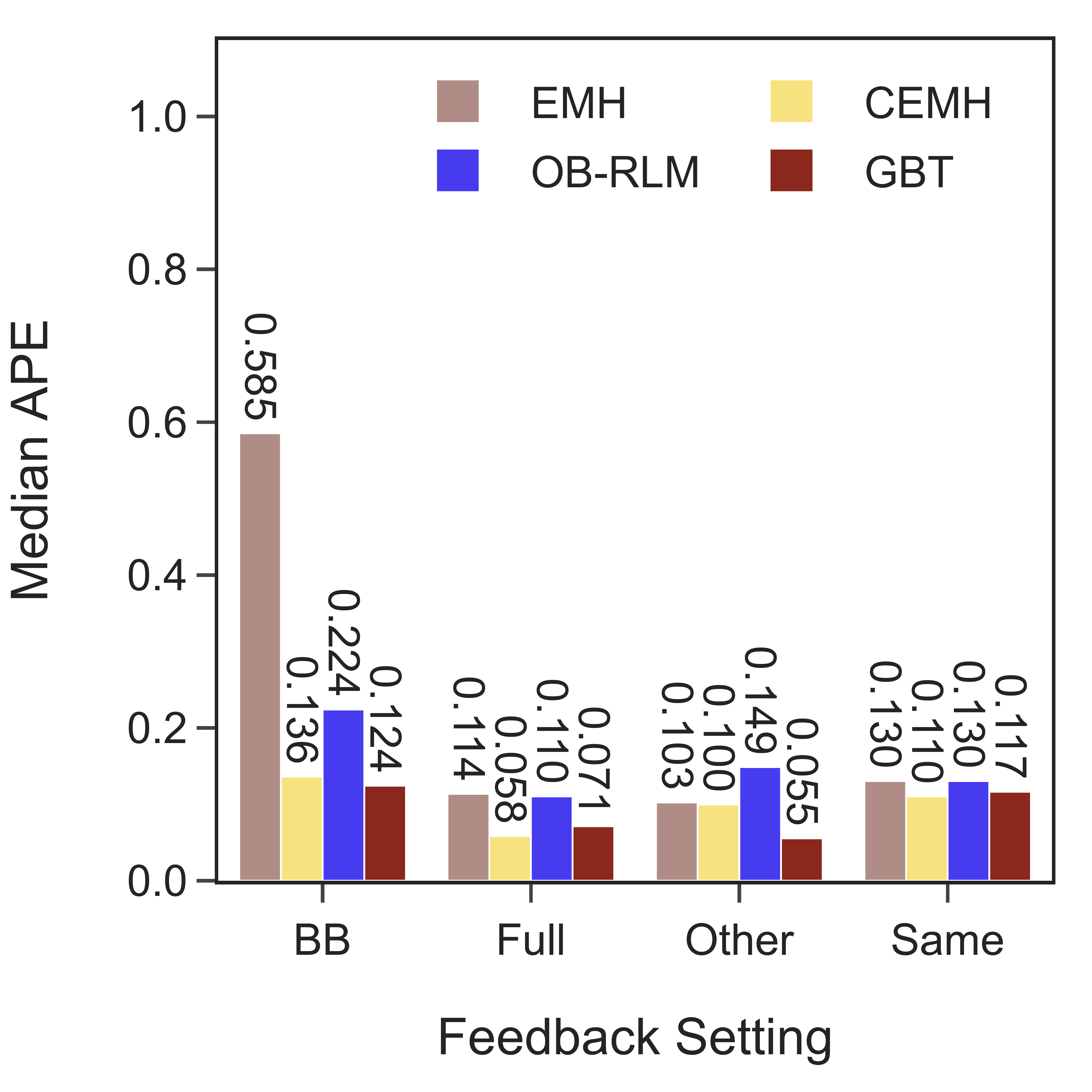}
        \label{fig:ae:setting:ape:1}
    }
    \caption{\emph{Allocative efficiency prediction error before and after initial price realizations as a function of feedback.} Comparison of MAPE values for AE for the first round for different feedback settings and price rules. Lower Median APE values denote better performance.}
    \label{fig:ae:setting:ape}
\end{figure}

\noindent Next, we evaluate the models predicting AE. The results in terms of observed deals and rounds are presented in \cref{tab:ae_ape}.
When rounds increase and players become more experienced, realized prices and all models become more effective at predicting the CEP, reaching a median APE below 7\%.
We notice that GBT outperforms median test APE except for the later rounds before a realized price is observed, where CEMH performs better.
In this case, observables that capture the system state are expected to be effective linear predictors of the system state, which, in our case, is the equilibrium state.
We check the statistical significance of our test results (see Appendix \cref{tab:ae_ape:wilcoxon}) by performing a non-parametric paired Wilcoxon test for all test samples across different predictive models.
Looking at different market sizes when evaluating AE MAPE performance in %\cref{tab:ae_ape:players}
the right side of \cref{tab:ae_ape}, we observe that GBT and CEMH perform better than other models in most cases, regardless of market size or number of observed deal prices.
The next result analysis evaluates our baselines on different information feedback settings during the first round~\cref{fig:ae:setting:ape}.
After the first deal happens, linear models, especially OB-RLM, perform well across most information feedback settings and price rules.
\rev{A spatial breakdown of the per-game best model across buyer and seller valuation-decile distances (\cref{sec:appendix:voronoi}) shows that non-linear models dominate in the first round for markets with lower-than-median bid and ask prices, a region that also contains the largest share of the dataset; linear models catch up in later rounds once realised prices accumulate.}

\subsubsection{EMH}
We observe that the EMH model may achieve lower-than-expected performance when results are aggregated at a time level, e.g., in \cref{tab:ae_ape} or \cref{fig:ae:setting:ape}, whereas it achieves around 25\% when results are aggregated at round level, e.g. in the right side
\cref{tab:ae_ape}.
In general, there are around 20 markets with first rounds with less than 60\% AE and very few or even no deals, mainly in Black Box settings.
During these rounds, several actions (bids/asks) are taken, and each test prediction is performed per action, shifting the method's median performance to lower values.
Based on \cref{fig:ae:setting:ape}, we see that all models achieve lower performance during the least efficient feedback setting Black Box (BB) and especially prior to no observed deals, 
as a closer evaluation shows that almost 46\% of the timesteps in BB are below 60\% AE.
Given that BB is one of the most common settings according to \cref{fig:sunburst} and that the train-tests split is done on half of the games of each treatment, we expect a high Median-APE when a lot of inefficient BB settings are sampled for the test sets.
The EMH only `predicts' well when the underlying market is, in fact, (close to) efficient, which is typically not entirely the case, especially in Black Box.

\subsubsection{CEMH} 
Since markets converge to being close to efficient over time, the EMH `improves' as a prediction over time and in later rounds. Correcting the EMH by the `known' inefficiency (as evaluated by group statistics on the aggregate data by market type) allows us to improve the prediction of a CEMH for early rounds.
The high performance of calculating group statistics across feeback settings and price rules, indicates that the information feedback plays a very important role in market efficiency.
This type of prediction overall does not become the best-performing prediction, except for the case of small and repeated markets but before price realization in that round. Indeed, the CEMH quite often outperforms the linear regressions in other situations too\footnote{This is one of the results that motivates us, as we shall discuss in the conclusion, to dig deeper into model-based/theory-informed prediction techniques going forward.}. 

\begin{figure}[htb]
    \centering
    \subfloat{
    \includegraphics[width=0.38\textwidth]{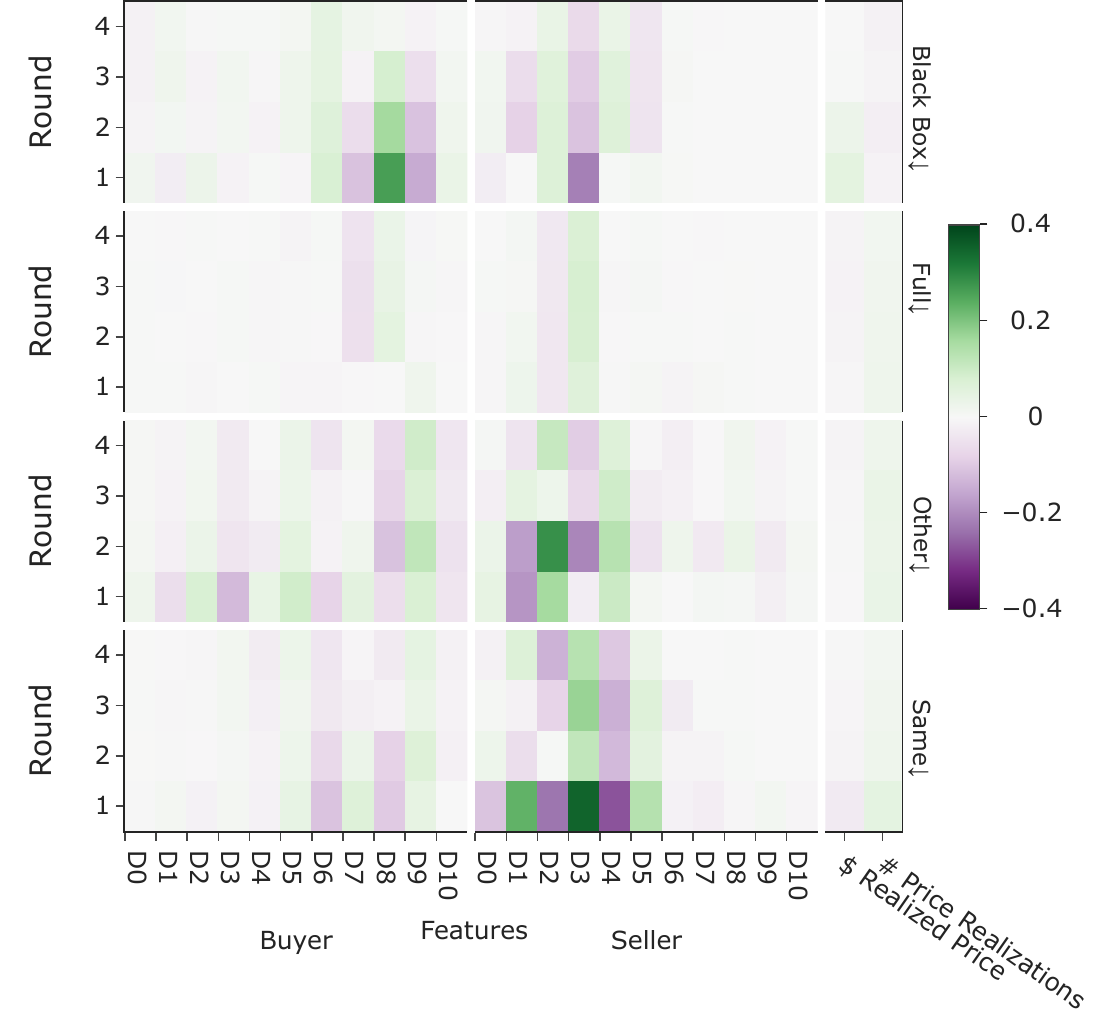}
    }
    \hfill
    \subfloat{
        \includegraphics[width=.52\textwidth]{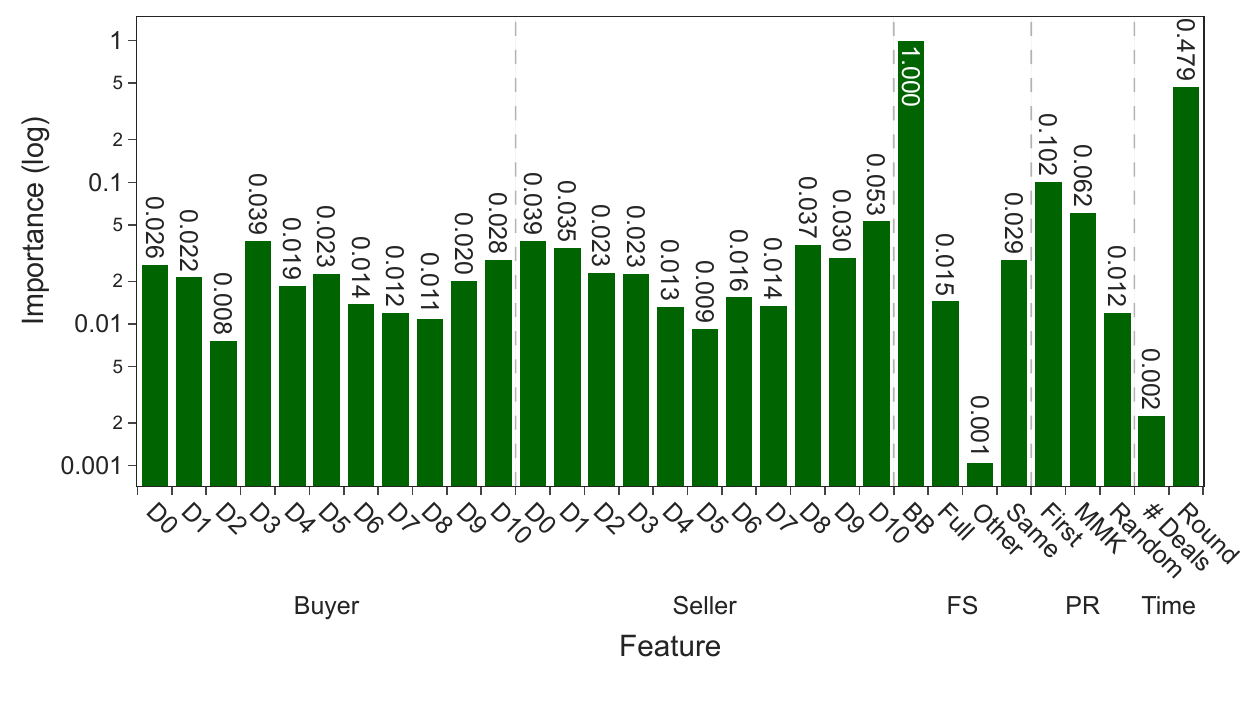}
    }     
    \caption{\emph{Comparison of OB-RLM coefficients (left) and GBT feature importance values (right) across different feedback setting groups.} We observe that the model assigns different coefficient values for different feedback settings. Intercept values are the highest (omitted for visual clarity), starting from 0.72 at round 1 and increasing to 0.84 at round 4, indicating that the model inputs do not contain enough information to predict AE across settings linearly. For GBT we observe that ``BB'' feedback setting and the ``round'' features have the highest importance for predicting AE. In both models we see that focusing in higher bid and lower ask deciles aids predictability of AE considerably.}
    \label{fig:ae:obrlm:params}
\end{figure}

\subsubsection{OB-RLM}
Initial result analysis on the training set of the first train-test split sample indicated that using feedback setting and price rule information would degrade the RLM performance. 
Therefore, we used only quantiles of players' bids and asks, deal prices and total deals as inputs.
We next look at the model's most influential and statistically significant coefficients in the left side of \cref{fig:ae:obrlm:params}. 
Higher intercept terms indicate the existence of non-captured information by the model inputs. OB-RLM does not perform particularly well, even with these adjustments, compared with the simpler versions used for CEP prediction in \cref{appdx:section:cep}. This means that AE seems to be harder to predict with simpler linear models on experimental data in this case. In fact, OB-RLM is quite often outperformed by the CEMH model, which does not use any orderbook information, not even realized prices.

\subsubsection{GBT}
For gradient boosting, we illustrate the mean log feature importance the right side of \cref{fig:ae:obrlm:params} across different training and test split samples.
Notably, the GBT model assigns higher importance to Black Box feedback and the round identifier.
This is expected, as Black Box markets tend to be less efficient~\citep{ikicacompeq}.
As regards the bid/ask deciles, lower quantiles of seller asks are the most important features of this model for AE prediction. Overall, GBT is the most predictively successful model, especially prior to price realization in the very first round.

\subsection{CEP Prediction Performance and Analysis}\label{sec:cep:pred:analysis}

\begin{table}[htb!]
    \caption{\emph{Left: CE price prediction error before and after initial price realizations.} CEP test set performance for per-round CEP values calculated separately before any price realization and once at least one price realized. Bold is better. GBT outperforms the other models when no deal price has yet been observed in the round. Otherwise, OB-RLM is the best-performing method.
    \emph{Right: CE price prediction error for small and large markets.} CEP test set performance for per-round CEP values calculated for `Small' markets ($<15$ players) and `Large' markets ($\geq$ 15) separately (market size 15 is the median market size during first round of a game.). Bold is better.
    }
    \centering
    \hrule
    \footnotesize
    \bgroup
    \def\arraystretch{1.0}
    \setlength\tabcolsep{3pt}
    \begin{tabular}{c || c | cccc | c | cccc}
        & \multicolumn{5}{c|}{\textbf{Price Realizations Analysis (Left)}} & \multicolumn{5}{c}{\textbf{Market Size Analysis (Right)}} \\
        Round & Price realizations & EMH & CEMH & OB-RLM & GBT & Market size & EMH & CEMH & OB-RLM & GBT \\
        \midrule
        \multirow[c]{2}{*}{1} & 0 & n/a & n/a & 0.191 & \B 0.135 & Small & 0.152 & 0.124 & 0.091 & \B 0.079 \\
        & 1+ & 0.109 & 0.092 & \B 0.061 & 0.077 & Large & 0.131 & 0.110 & \B 0.063 & 0.084 \\
        \cline{1-6} \cline{7-11}
        \multirow[c]{2}{*}{2+} & 0 & n/a & n/a & 0.109 & \B 0.099 & Small & 0.094 & 0.083 & 0.072 & \B 0.070 \\
        & 1+ & 0.062 & 0.055 & \B 0.048 & 0.051 & Large & 0.071 & 0.064 & \B 0.047 & 0.051 \\
    \end{tabular}
    \egroup
    \label{tab:cep_ape}
    \hrule
\end{table}

\begin{figure}[htb]
    \subfloat[Prior to any price realization]{
        \includegraphics[width=0.45\textwidth]{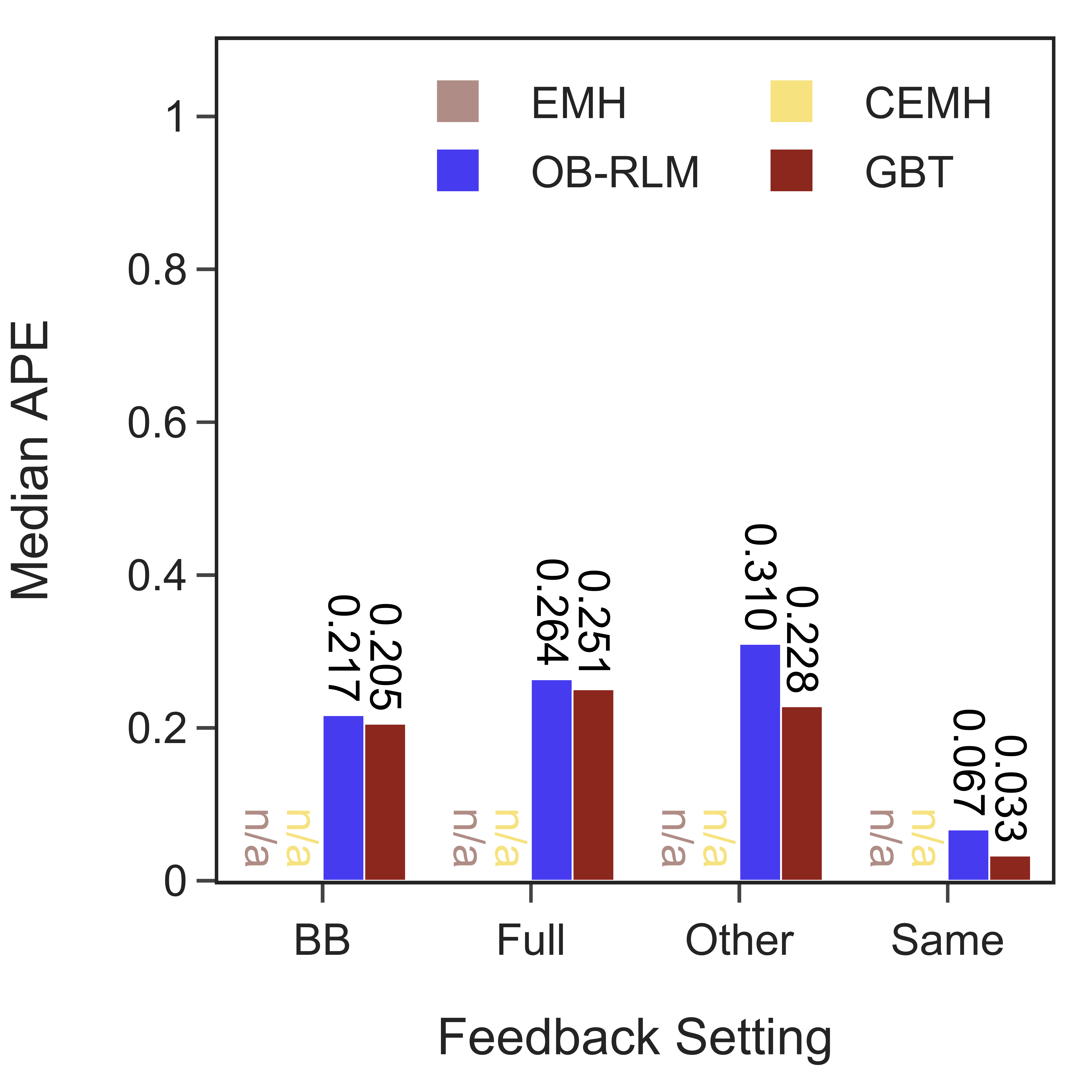}
        \label{fig:cep:setting:ape:0}
        }
    \hfill
    \subfloat[At least one price realization]{
        \includegraphics[width=0.45\textwidth]{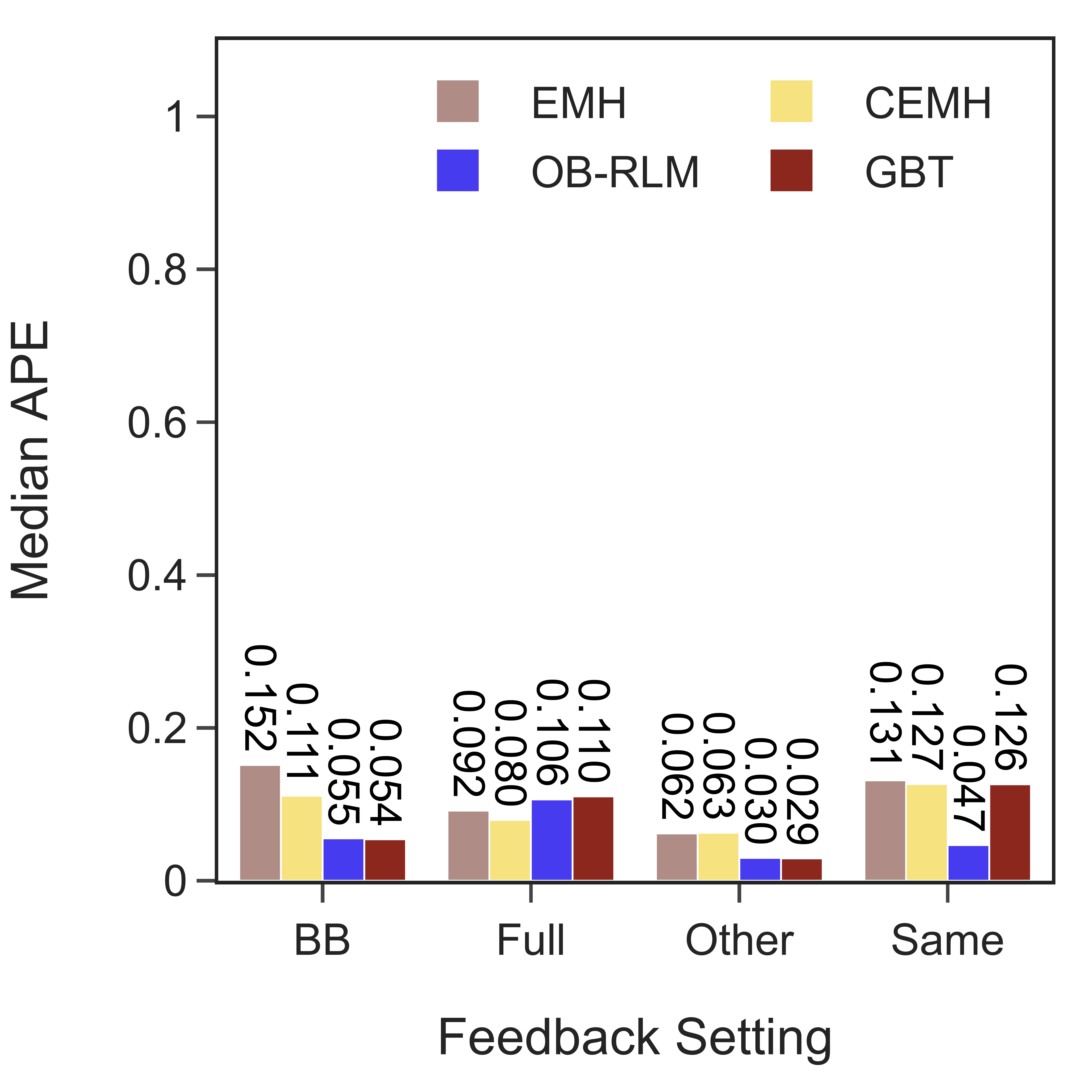}
        \label{fig:cep:setting:ape:1}
        }
    \caption{\emph{CE price prediction error before and after initial price realizations as a function of feedback.} Comparison of MAPE values for CEP for the first round for different feedback settings. Lower Median APE values denote better performance.}
    \label{fig:cep:setting:ape}
\end{figure}

\noindent First, we evaluate the models predicting CEP. The results in terms of observed deals and rounds are presented in~\cref{tab:cep_ape}.
As expected from previous work~\citep{ikicacompeq}, CEP is more predictable by linear models as more realized prices appear within the same round.
When rounds increase and players become more experienced, realized prices and all relevant models become more effective at predicting the CEP, reaching a median APE below 5\%.
In particular, when we look at the high and low boundaries of the CEP interval~\citep{ikicacompeq}, we notice that the interval gap is around 2\% wide in terms of APE around the CEP, showcasing that all evaluated models can predict CEP accurately --- near that gap value.
We notice that GBT outperforms in terms of median test APE prior to the observation of any realized price, 
which is expected as, at that time, the market is in a transient part before the competitive marketing equilibrium.
In physics, systems near an equilibrium point can be predicted by applying a linearization (e.g., first-order Taylor expansion) to the vector field of the dynamics~\citep{strogatz2018nonlinear}. 
In this case, observables that capture the system state can be expected to be effective linear predictors of the system state; in our case, this should resemble the competitive equilibrium price.
When away from the equilibrium, transient dynamics are often predicted by non-linear models~\citep{strogatz2018nonlinear} ---technically, they would require higher order terms in the Taylor expansion--- to the dynamics state.
In this case, we expect that non-linear complex models can predict such phenomena better, as the results of  \cref{tab:cep_ape} empirically point out. 
We also check the statistical significance of our test results (see \cref{tab:ce_ape:wilcoxon}) by performing a non-parametric paired Wilcoxon test for all test samples across different predictive models.
Looking at different market sizes when evaluating CEP MAPE performance in the right side of \cref{tab:cep_ape}
% \cref{tab:cep_ape:players}
, we observe that the non-linear GBT performs better than other models in most cases, especially prior to any price realization.
Notably, the main drop in performance of the GBT model is during the first round for larger markets and once more deal prices appear. \rev{The same valuation-decile spatial breakdown for CEP (\cref{sec:appendix:voronoi}) shows GBT outperforming the linear models before any deal is realised, with linear models catching up once prices accumulate.}
The final result analysis evaluates our baselines on different information feedback settings during the first round~\cref{fig:cep:setting:ape}.
Before any deal price is observed, GBT outperforms the OB-RLM in all cases, while the OB-RLM model outperforms GBT in Black Box and Full feedback settings after the first deal occurs.

\subsubsection{EMH and CEMH}
Both models are unable to predict prior to any price realization. Once prices realize and become available for prediction, the EMH is off by the amount that prices are short (below) of CEP, and the CEMH corrects these gaps by best estimators (see \cref{tab:cep:cemh:coeffs} for the overall correction.

\begin{table}[htb]
    \caption{\emph{Price correction by market type.} Median Values for CEMH Coefficients across different feedback settings and price rules. All coefficient values are statistically significant. E.g., a coefficient of 1.05 indicates that the average price realization should be increased by five percent to obtain the best predictor for CE price. Note that, at the median value, the realized price coefficients often scale the realized prices to values that are around 3-8\% higher. 
The scaling is stronger at Black Box feedback settings and especially for MMK and Random price rules.}
    \centering
        \begin{tabularx}{\textwidth}{l c | X c}
        \toprule
         & Price correction & Price rule & Feedback \\
        \midrule
         & 1.05 & First & Black Box \\
         & 1.08 & MMK & Black Box \\
         & 1.06 & Random & Black Box \\
         & 1.03 & First & Full \\
         & 1.04 & First & Other \\
         & 1.02 & First & Same \\
         & 1.03 & MMK & Same \\
        \bottomrule
        \end{tabularx}
        \label{tab:cep:cemh:coeffs}
\end{table}

\subsubsection{OB-RLM}
\begin{figure}[htb]
    \centering
    %\begin{subfloat}[T]{0.98\textwidth}
        \includegraphics[width=0.88\textwidth]{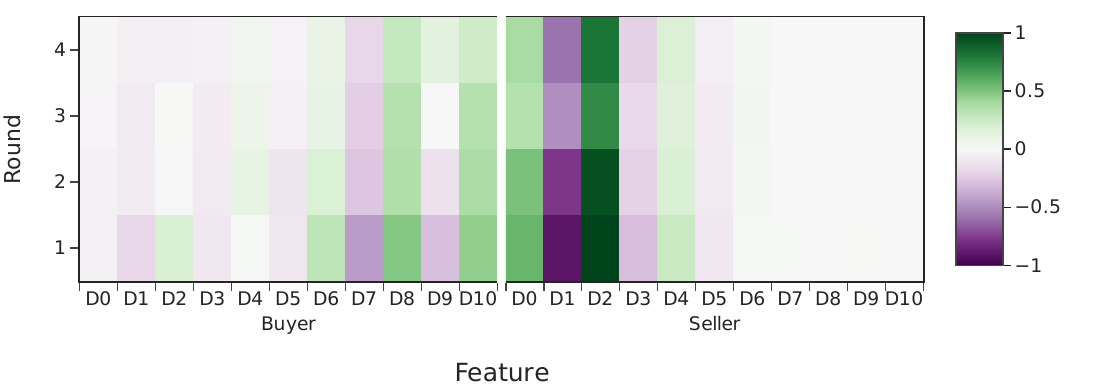}
        %\caption{}
    %\end{subfloat}
    \caption{\emph{Mean coefficient values of OB-RLM for CE price prediction of buyers' bid and sellers' ask deciles.} Coefficient values for input features of OB-RLM model.}
    \label{fig:ce_ape:obrlm:params}
\end{figure}

Initial result analysis on the training set of the first train-test split sample indicated that using feedback setting and price rule information would degrade the RLM performance. 
Therefore, we used only the deciles of players' bids and asks as input features to the model.
We next look at the statistically significant coefficients of the model in \cref{fig:ce_ape:obrlm:params}, observing that higher bid deciles and lower ask deciles have a stronger influence on the prediction of CEP.
This model indicates that buyer bids lose predictive power as rounds progress.\footnote{This is consistent with a finding from~\citet{ikicacompeq}, whereby buyer bids are more aggressive and influential earlier in the round but then become less relevant as buyers tend to yield faster over time.} The OB-RLM is overall the best-performing predictor of CEP, except for small markets and predictions prior to any price realization.

\subsubsection{GBT}
\begin{figure}[htb]
        \centering
        \includegraphics[width=.88\linewidth]
        {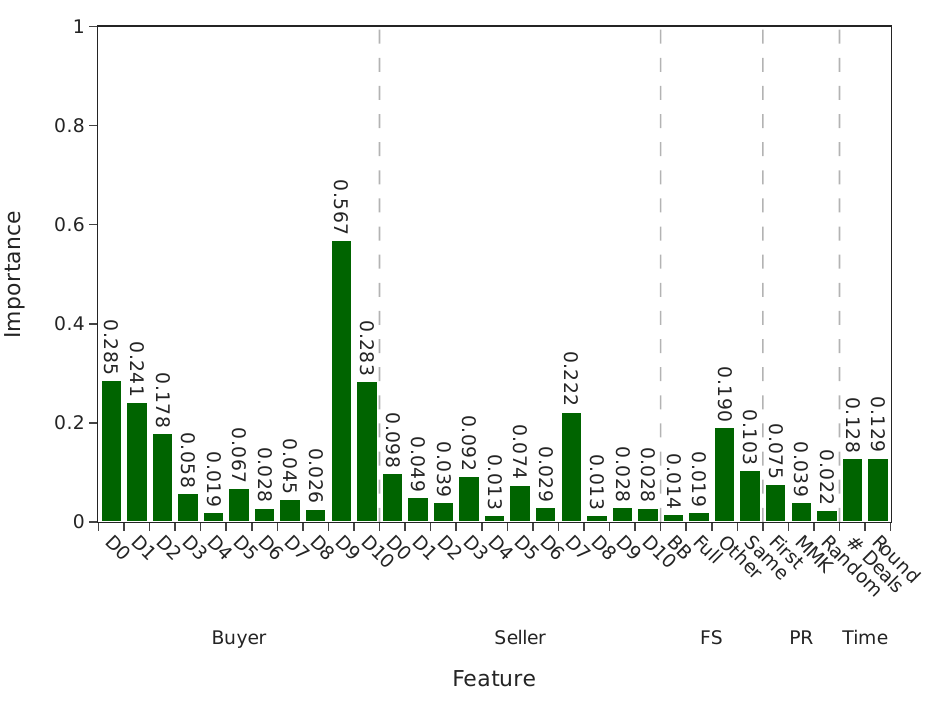}
        \caption{\emph{Relative relevance of input features of GBT for CE price.} Mean relative importance values for all input features provided to GBT, including buyers' bid deciles, sellers' ask deciles, feedback type, price rule, and timing.}
        \label{fig:ce_ape:gbt:importance}
\end{figure}

For gradient boosting, we illustrate the mean feature importance in \cref{fig:ce_ape:gbt:importance} (LHS) across different training and test split samples.
Notably, the GBT model assigns higher importance to higher and lower buyer quantiles.
While the inclusion of feedback settings and price rules degrades the performance of the linear model, GBT assigns high importance to specific feedback settings and price rules. The OB-RLM outperforms the GBT model in many situations but is the best-performing model generally prior to any price realizations and for small markets in the early rounds. %

\subsection{\rev{Diagnostics and feature response}}\label{sec:diagnostics}

\rev{We complement the bucket-aggregated APE in \cref{tab:ae_ape,tab:cep_ape} with per-sample diagnostics for the two orderbook-based models (OB-RLM, GBT) and a feature-response analysis for GBT.
The other models are linear maps or identity maps of inputs already interpreted elsewhere (\cref{fig:ae:obrlm:params,fig:ce_ape:obrlm:params,tab:cep:cemh:coeffs}), so they are not repeated.
All panels use the same held-out test split ($\mathit{sample\_id} = 0$, $N = 9111$ rows) and the same four $(\vround, n)$ buckets as \cref{tab:ae_ape,tab:cep_ape}, where $n$ is the number of deals observed so far in the round.}

\begin{figure}[htb]
\centering
\includegraphics[width=.95\textwidth]{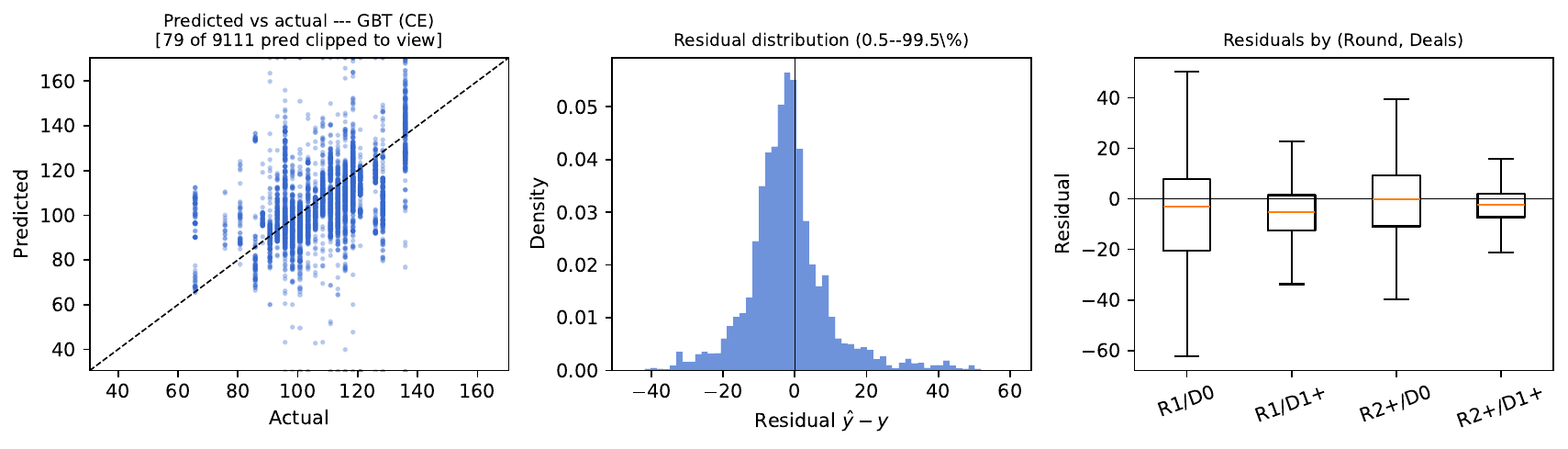}\\
\includegraphics[width=.95\textwidth]{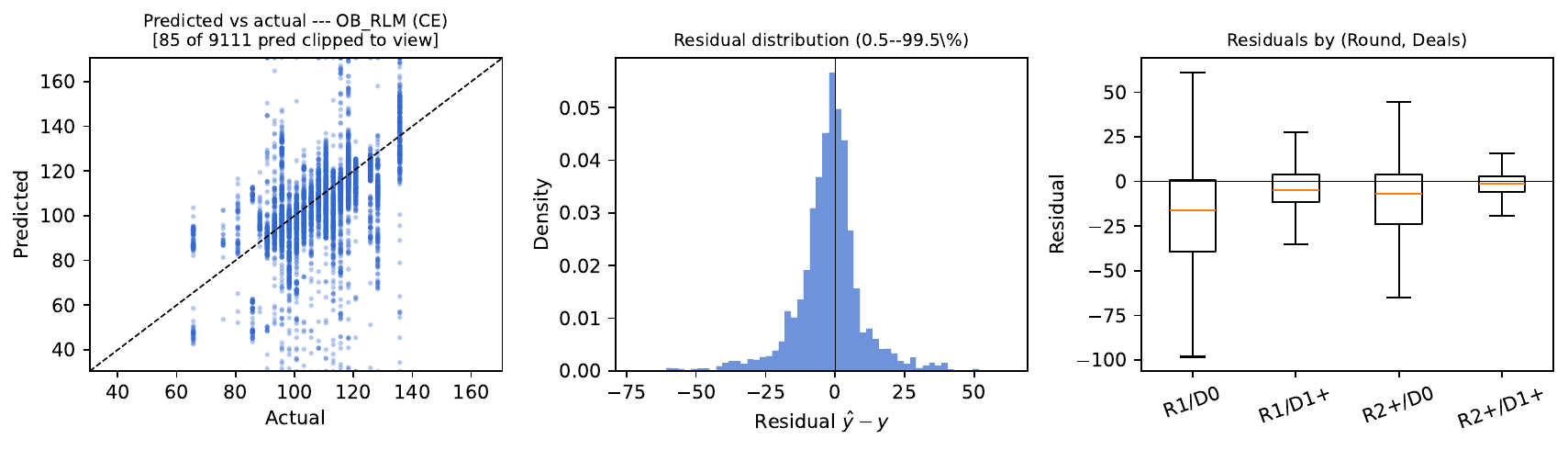}
\caption{\emph{\rev{M2 diagnostics for CEP.}} \rev{Top: GBT. Bottom: OB-RLM. Left: predicted vs.\ actual; both axes are clipped to $[30.5, 170.5]$ and the count of clipped points is annotated above the panel. Middle: residual distribution restricted to the $0.5$--$99.5\%$ quantile range. Right: residual boxplot ($\hat{p} - p$, outliers suppressed) across the four main-text buckets. The per-row IQR-driven extremes that are clipped on the left panel are visible in the upper tail of the $(\vround{\geq}2, n{=}0)$ box on the right.}}
\label{fig:diagnostics:m2:cep}
\end{figure}

\begin{figure}[htb]
\centering
\includegraphics[width=.95\textwidth]{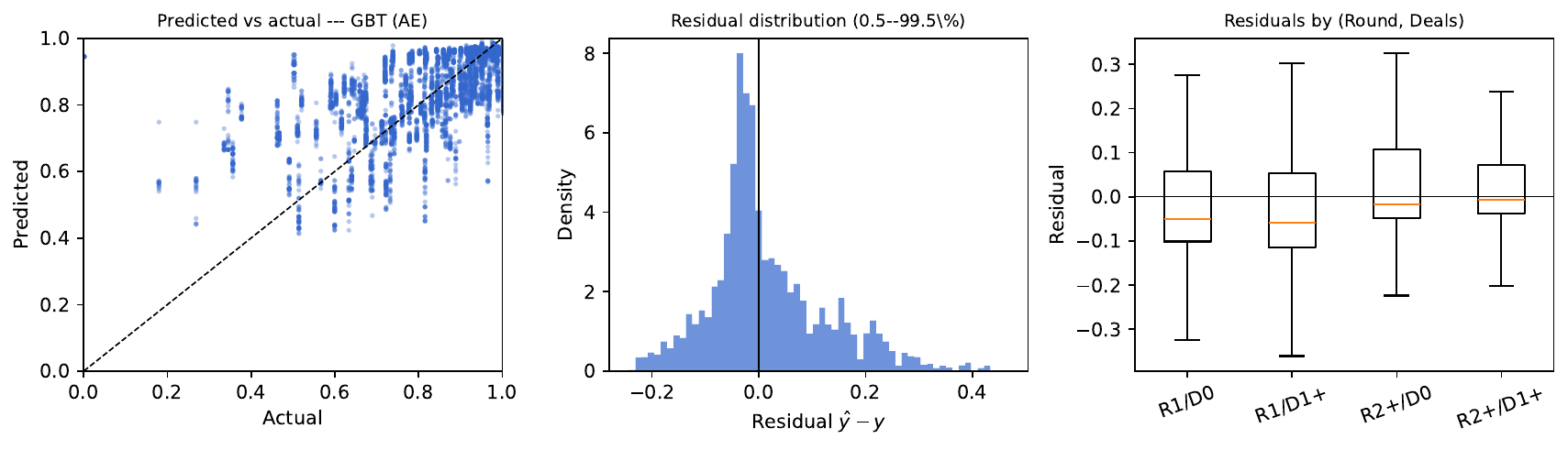}\\
\includegraphics[width=.95\textwidth]{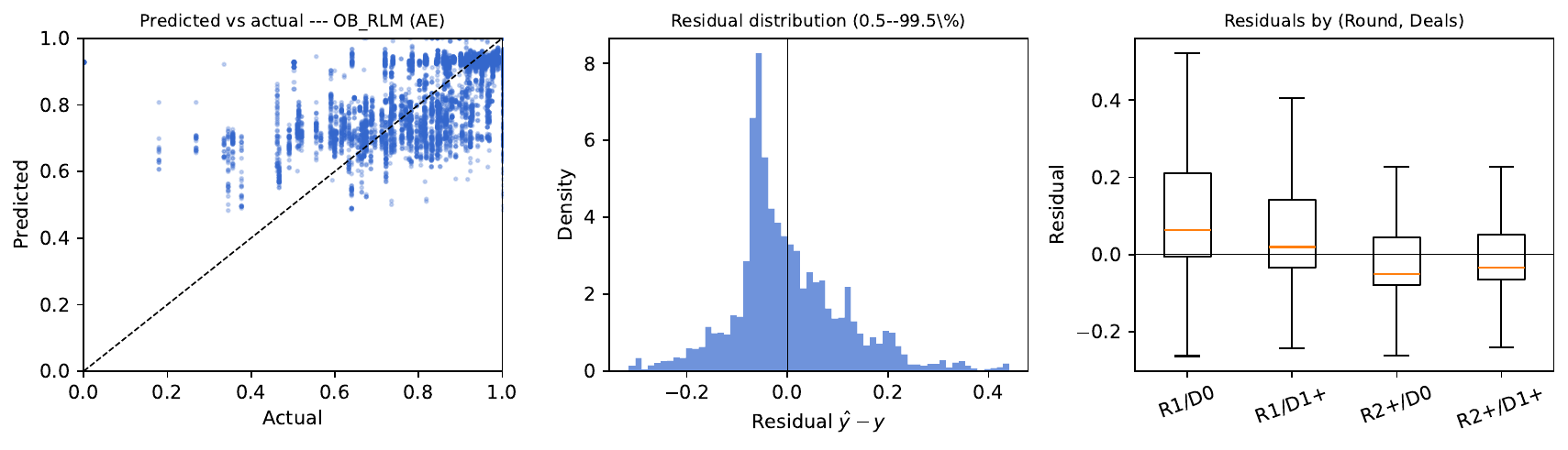}
\caption{\emph{\rev{M2 diagnostics for AE.}} \rev{As in \cref{fig:diagnostics:m2:cep} but for the AE task; no scatter clipping is needed since $E$ is bounded in $[0,1]$.}}
\label{fig:diagnostics:m2:ae}
\end{figure}

\paragraph{\rev{Per-bucket residuals (\cref{fig:diagnostics:m2:ae,fig:diagnostics:m2:cep}).}}
\rev{Both models exhibit clear residual heteroskedasticity across buckets, in line with the APE differences reported in the main text.
For AE, the GBT median APE drops from $0.144$ in $(\vround{=}1, n{=}0)$ to $0.051$ in $(\vround{\geq}2, n{\geq}1)$ and the residual standard deviation halves between the two ($0.18 \to 0.11$); the bias is small in absolute value but switches sign across buckets ($+1.3{\times}10^{-3}$ in $(\vround{=}1, n{=}0)$, $-2.1{\times}10^{-2}$ in $(\vround{=}1, n{\geq}1)$, $+5.4{\times}10^{-2}$ and $+2.2{\times}10^{-2}$ in the $\vround{\geq}2$ rows).
The OB-RLM AE residuals share the same bucket ordering but carry a substantially larger positive bias of $+0.114$ in $(\vround{=}1, n{=}0)$, consistent with the linear estimator defaulting toward an average efficiency level when no orderbook information has yet accumulated.
For CEP, the GBT median APE rises from $0.047$ in $(\vround{\geq}2, n{\geq}1)$ to $0.165$ in $(\vround{=}1, n{=}0)$.
The RMSE in $(\vround{\geq}2, n{=}0)$ is inflated to $416$ price units, but the median APE in the same bucket is $0.102$ and the $99$th percentile of $|\hat{p}-p|$ is $120$, so the RMSE is dominated by a small number of denormalisation extremes rather than by a systematic regression failure.\footnote{The orderbook features are normalised per row by the IQR of the $22$ entries of the concatenated decile vector $X_q = \begin{smallmatrix} \vquantilevec{\vbid} & \vquantilevec{\vask}\end{smallmatrix}$ defined in \cref{sec:dataset}. When all $22$ entries collapse onto a single point the IQR is replaced by $1$; conversely, when one row has $\vquantile{\vbid}{0}$ near zero and $\vquantile{\vask}{1}$ near the upper allowed limit (e.g.\ before any deal in a wide-allowed-bid treatment), the per-row IQR can reach several thousand and a modest normalised prediction is amplified by the same factor on the way back to raw price units. The predicted-vs-actual scatter in \cref{fig:diagnostics:m2:cep} clips $79/9111$ (GBT) and $85/9111$ (OB-RLM) such points to keep the bulk of the plot legible; the count is annotated above the panel.}
The same per-row IQR mechanism produces the heavy upper tail visible in the OB-RLM CEP residual histogram in $(\vround{\geq}2, n{=}0)$.}

\begin{figure}[htb]
\centering
\includegraphics[width=\textwidth]{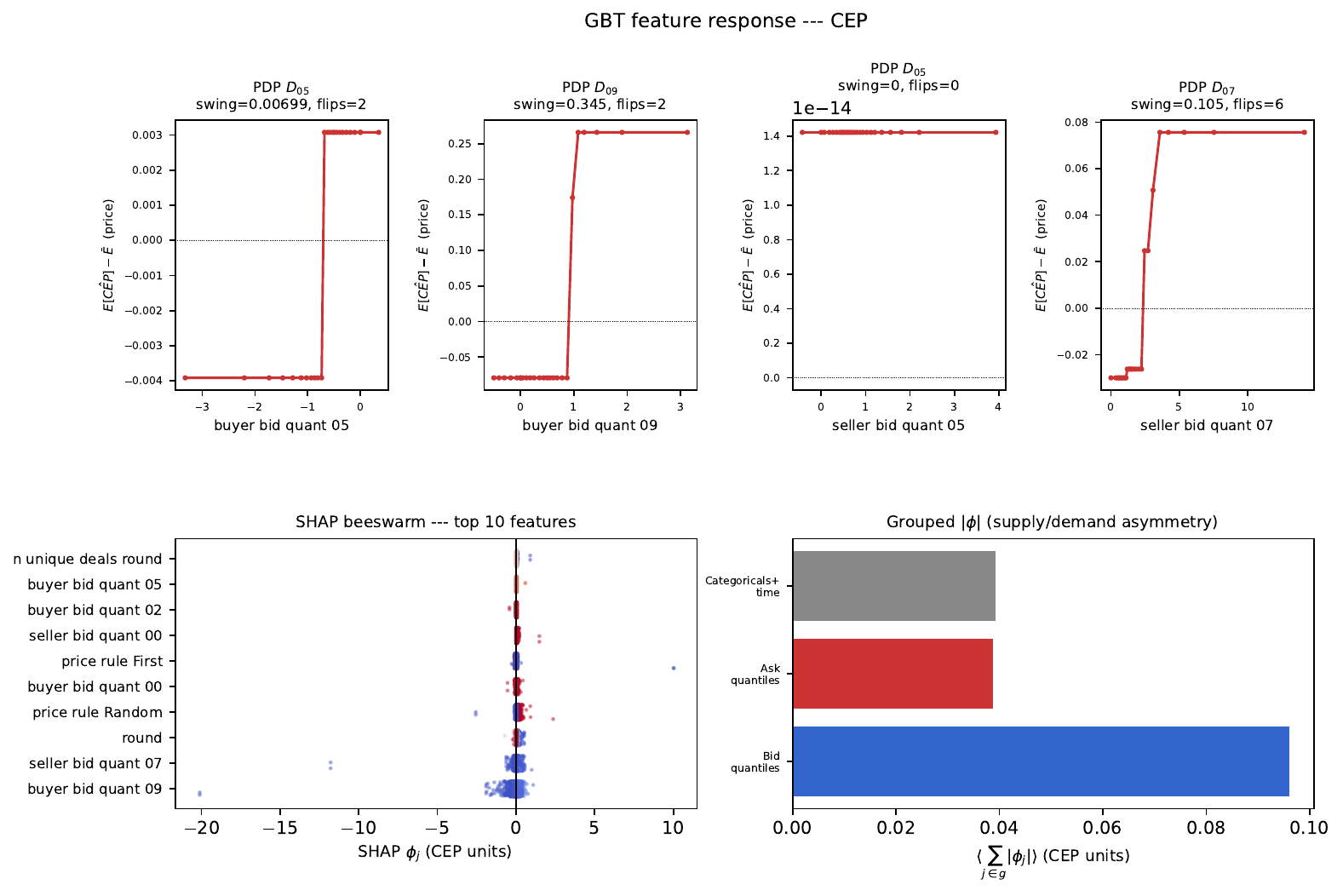}
\caption{\emph{\rev{GBT feature response for CEP.}} \rev{Top row: PDPs centred on $\bar{E}[\hat{p}]$ for the bid and ask medians $\vquantile{\vbid}{0.5}, \vquantile{\vask}{0.5}$ and for the highest-importance bid and ask extremes $\vquantile{\vbid}{0.9}, \vquantile{\vask}{0.7}$; the $y$-axis is in raw price units, panel title reports the $D_5$--$D_{95}$ swing of the curve and the number of sign flips of $\partial \hat{p}/\partial \vquantile{\cdot}{\pi}$. Bottom-left: SHAP beeswarm for the ten highest mean-$|\phi_j|$ features; colour encodes the feature value from low (blue) to high (red). Bottom-right: grouped $\langle \sum_{j \in g} |\phi_j| \rangle$ for bid quantiles, ask quantiles and categoricals/time. Per-row IQR is applied to denormalise SHAP and PDP back to raw price units before averaging, so all amplitudes are directly comparable to the realised price.}}
\label{fig:diagnostics:r29:cep}
\end{figure}

\begin{figure}[htb]
\centering
\includegraphics[width=\textwidth]{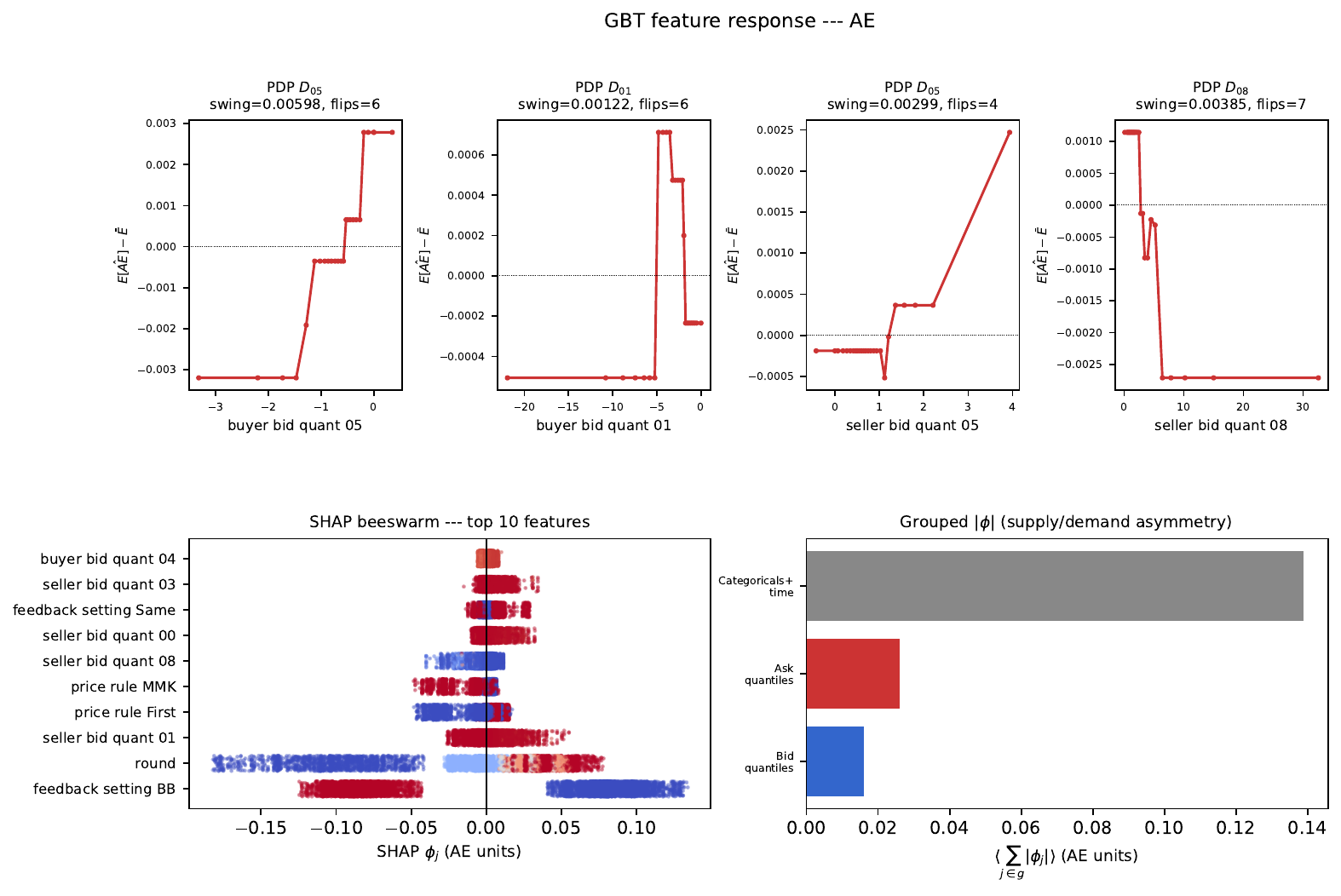}
\caption{\emph{\rev{GBT feature response for AE.}} \rev{As in \cref{fig:diagnostics:r29:cep} but for the AE task; the SHAP and PDP $y$-axes are in raw $E$ units in $[0,1]$.}}
\label{fig:diagnostics:r29:ae}
\end{figure}

\paragraph{\rev{Feature attributions (\cref{fig:diagnostics:r29:ae,fig:diagnostics:r29:cep}).}}
\rev{For each task we report (i) PDPs centred on the grand-mean prediction $\bar{E}[\hat{y}]$ for the bid and ask medians $\vquantile{\vbid}{0.5}, \vquantile{\vask}{0.5}$ and for the highest-importance bid and ask extremes; (ii) a SHAP beeswarm of the top ten features by mean $|\phi_j|$ computed with TreeSHAP on the fitted CatBoost model; and (iii) a grouped attribution $\langle\sum_{j\in g}|\phi_j|\rangle$ for $g \in \{\text{bid quantiles}, \text{ask quantiles}, \text{categoricals/time}\}$.
SHAP and PDP $y$-axes are reported in raw target units, AE in $[0,1]$, CEP in price units, so the absolute response amplitude is directly readable.}

\rev{For AE, $\mathit{feedback\_setting}{=}\text{BB}$ and $\vround$ dominate ($\langle|\phi|\rangle \approx 0.083$ and $0.040$, i.e.\ $8.3$ and $4.0$ percentage points of efficiency), and the highest-ranked quantile feature is $\vquantile{\vask}{0.1}$ at $0.007$.
The grouped attribution gives $0.139$ for categoricals/time vs.\ $0.026$ for the seller-ask deciles vs.\ $0.016$ for the buyer-bid deciles; once the categorical effect is conditioned out, the ask side carries a factor $\approx 1.6$ more attribution mass than the bid side.
The four PDPs are flat in absolute terms (swing $\le 0.006$, i.e.\ $\le 0.6$ percentage points of $E$) and non-monotone, with between $4$ and $7$ sign flips of $\partial \hat{E}/\partial \vquantile{\cdot}{\pi}$ along the empirical $2$nd--$98$th percentile sweep, consistent with the categoricals carrying most of the explanatory weight and the residual quantile attribution being smeared across a near-collinear set of inputs.}

\rev{For CEP, $\vquantile{\vbid}{0.9}$ and $\vquantile{\vask}{0.7}$ are the two leading features ($\langle|\phi|\rangle \approx 0.075$ and $0.031$ price units), followed by $\vround$ and the price-rule dummies at $\sim 10^{-2}$.
The grouped attribution reverses the AE asymmetry: the buyer-bid deciles carry $0.096$ vs.\ $0.039$ for the seller-ask deciles (ratio $\approx 2.5$).
The PDP sweep at $\vquantile{\vbid}{0.9}$ on the buyer side spans $0.34$ price units ($\approx 0.3\%$ of $\bar{p} \approx 108.7$), monotone in the relevant direction; the bid and ask medians $\vquantile{\vbid}{0.5}, \vquantile{\vask}{0.5}$ are essentially flat (swing below $10^{-2}$ price units, $\sim 0.01\%$); the ask-side $\vquantile{\vask}{0.7}$ curve has a swing of $0.11$ price units with $6$ sign flips.
The non-monotonicity at $\vquantile{\vask}{0.7}$ is consistent with TreeSHAP's credit splitting on near-collinear quantile features, which is precisely the failure mode the grouped $|\phi|$ summary is designed to absorb; we therefore use the grouped value as the headline quantitative read of the bid/ask asymmetry and treat the per-feature PDP swings only as response-shape diagnostics.}

\paragraph{\rev{Caveats.}}
\rev{The orderbook deciles satisfy $\vquantile{\cdot}{\pi_1} \leq \vquantile{\cdot}{\pi_2}$ for $\pi_1 \leq \pi_2$ by construction, so the PDPs should be read as response functions of the fitted model, not as causal marginal effects.
The complementary TreeSHAP attributions stay on the observed manifold; the bid- and ask-grouped sums in the rightmost panel of \cref{fig:diagnostics:r29:ae,fig:diagnostics:r29:cep} absorb any within-group credit splitting on the near-collinear deciles.
PDPs are intentionally omitted for categorical features: a PDP over $k$ levels reduces to the $k$ bucket means already reported in \cref{fig:ae:setting:ape,fig:cep:setting:ape,tab:ae_ape,tab:cep_ape}, while the within-level dispersion that the beeswarm exposes is not recoverable from a level-mean plot.}

\subsection{\rev{Input-feature ablations}}\label{sec:ablation:discussion}

\rev{We probe the contribution of the two most informative input categories on top of the running orderbook quantiles: the treatment descriptors plus round and $n$, and the realised deal price. The full detail tables and implementation are reported in the appendix (\cref{sec:appendix:ablations}). We summarise the main findings here.}

\rev{Removing the treatment descriptors and the round and $n$ features leaves the running bid and ask quantiles alone. GBT for CEP is numerically unchanged, so the quantiles already carry the predictive content that the paper attributes to the model; CEMH-global for CEP is also essentially unchanged because the per-treatment correction coefficients are close to one and replacing them with a single global scalar barely changes the predictions. On the AE side both OB-RLM and GBT lose accuracy, most visibly in the round-one no-deals bucket for GBT ($0.168$ to $0.316$) and in later rounds with at least one deal for OB-RLM ($0.093$ to $0.136$). The feedback-setting grouping and the $n$ term therefore carry information about proximity to clearing. Removing the realised deal price from OB-RLM for AE leaves the aggregate median APE essentially unchanged ($0.107$ to $0.106$) and keeps all four buckets within one hundredth of the original fit. The AE results of \cref{tab:ae_ape} are therefore not driven by circular reuse of the realised price.}

\subsection{Discussion: From Research to Real Markets}\label{sec:laboratory}

Our experimental results demonstrate robust efficiency prediction across diverse market conditions, yet a critical question remains: can these insights translate meaningfully to real-world trading environments? The gap between controlled laboratory settings and complex commercial markets presents both challenges and opportunities for practical implementation. Rather than dismissing this gap as insurmountable, we argue that it precisely motivates the need for machine learning approaches in inverse market analysis, where traditional economic theory alone proves insufficient to extract efficiency signals from observable market data.

\subsubsection{The Electronic Component Exchange: A Natural Laboratory}

Electronic component exchanges represent a compelling bridge between the experimental and commercial markets, exhibiting structural features that align remarkably with our framework, while remaining economically significant. 
Consider manufacturers sourcing standardized semiconductors, capacitors, and integrated circuits from multiple suppliers through digital platforms, 
a potential market processing billions of dollars annually with characteristics that mirror our experimental design.

These markets operate within distinct temporal boundaries that create the stability our models require. 
Weekly or monthly procurement cycles establish periods where buyers maintain relatively fixed production requirements driven by manufacturing forecasts, 
while suppliers work with predictable capacity costs and inventory positions. 
This temporal structure provides the valuation stability that enables our efficiency prediction framework, 
creating windows where our assumption of fixed reservation values becomes practically meaningful.

Repetition of interaction patterns in component exchanges distinguishes them fundamentally from one-shot procurement events.
Unlike traditional request-for-quote processes, where each transaction exists in isolation, 
these platforms facilitate ongoing relationships where the same participants trade regularly across multiple cycles.
This repetition generates the orderbook patterns and behavioral consistency that machine learning models require to identify efficiency signals embedded in observable bid and ask data. 
The algorithmic nature of our approach becomes essential here, because human analysts cannot process the complex, 
high-frequency patterns that emerge from repeated strategic interactions.
In this case, ML-based inference is not just useful, but practically necessary.

It is highly likely that information asymmetries in these markets create the exact problem our framework addresses. 
Buyers' true urgency levels, driven by inventory constraints and production deadlines, remain hidden from suppliers. 
Similarly, the actual production costs, capacity limitations, and strategic inventory positions of the sellers remain confidential. 
Traditional economic analysis may not be enough in this environment, because the fundamental inputs, namely true willingness to pay and accept, remain unobservable. 
Our machine learning approach is so promising precisely because it can extract efficiency signals from the complex patterns in observable orderbook data without requiring access to these private valuations.

\subsubsection{Confronting Reality: When Theory Meets Practice}

A more thorough assessment reveals significant divergences between our experimental assumptions and commercial market realities. 
Most B2B platforms operate through request-for-quote systems~\cite{schoenherr2008use} or sealed-bid formats~\cite{zeithammer2010sealed} rather than continuous double auctions, fundamentally altering information disclosure mechanisms. 
Component valuations fluctuate with demand forecasts and supply chain disruptions, violating our temporal stability assumptions. 
Real procurement decisions incorporate multiple dimensions beyond price/quality standards, delivery reliability, and supplier relationships that our single-attribute framework ignores. Professional traders employ sophisticated strategies, including inventory management and forward contracting, that exceed the strategic complexity of our experimental subjects.

However, these limitations support rather than undermine the case for machine learning approaches. 
Traditional economic models, built on strong assumptions about market mechanisms and participant behavior, 
become even less applicable when these assumptions fail. 
Machine learning models, in contrast, can generalize to various market structures and learn patterns that remain stable even when the underlying mechanisms differ. 
The very complexity that may render theoretical approaches less applicable is also the natural domain where data-driven methods excel. 
This complexity arises from multiple interacting strategic behaviors, non-stationary valuations, and multidimensional decision criteria.

Rather than forcing real markets to conform to experimental assumptions, we propose future adaptive implementation strategies that leverage our framework's core insights while acknowledging practical constraints. 
The key insight is that efficiency prediction provides value even when perfect accuracy remains elusive, i.e. directional indicators and relative comparisons often suffice for meaningful platform optimization. To achieve this, we require the following:

\begin{itemize}
    \item \textbf{Stable window analysis} represents our most immediate practical application. 
    By identifying periods where component valuations remain relatively constant, e.g. 2-4 hour trading sessions during stable market conditions, our models can provide meaningful efficiency predictions. 
    During these windows, the gap between our experimental assumptions and market reality narrows sufficiently for reliable inference.
    \item \textbf{Market health monitoring} provides a diagnostic value that complements rather than replaces human judgment.
    Our models can flag trading sessions with persistently low predicted efficiency, alerting platform operators to potential manipulation, technical issues, or structural problems. 
    By identifying participant segments that consistently contribute to inefficiency, platforms can target educational interventions or adjust market rules to improve overall performance. 
    The machine learning approach becomes essential here because human operators cannot monitor efficiency patterns across hundreds of simultaneous trading sessions with the consistency and scale provided by automated systems.
\end{itemize}

In summary, our proposed framework extends naturally to mechanism design applications where efficiency predictions inform platform optimization decisions. 
Marketplace operators can systematically compare auction formats, testing whether transitions between information disclosure regimes improve allocative outcomes. 
Dynamic fee structures can respond to predicted inefficiencies, with higher transaction costs applied during periods of anticipated poor performance to incentivize more careful bidding. 
These applications do not require perfect efficiency prediction, as they need reliable relative comparisons between different market configurations, a more achievable goal that remains valuable for platform design.

\subsubsection{The Case for Inverse Analysis: When Forward Modeling Fails}

Our framework's fundamental contribution lies in recognizing that traditional forward modeling, namely building theoretical predictions from assumed market structures, becomes inadequate when market complexity exceeds theoretical tractability. 
Real markets exhibit emerging behaviors, strategic interactions, and informational complexities that resist analytical solutions. 
The inverse approach, extracting efficiency signals from observable data using machine learning methods, provides the only practical path forward when theoretical models cannot fully capture market dynamics.
Machine learning models may identify subtle patterns in high-dimensional orderbook data that human analysts may miss, detect efficiency signals that emerge from the interaction of multiple strategic behaviors, and adapt to changing market conditions without requiring explicit theoretical understanding of underlying mechanisms.

The experimental foundation ensures that our predictions maintain theoretical grounding while remaining computationally tractable for practical deployment. 
By validating our approach in controlled settings where ground-truth efficiency measures exist, we establish confidence in the method's core principles before extending to environments where such validation becomes impossible. 
This progression from laboratory to market represents not a compromise, but a natural evolution of economic methodology toward problems that require computational rather than purely analytical solutions.

Future validation through field experiments in real B2B markets will refine our understanding of when and how these tools perform optimally, 
but the fundamental case for inverse efficiency analysis remains compelling wherever market complexity exceeds the bounds of traditional economic theory. 
Our framework provides principled efficiency metrics where none currently seems to exist, enabling systematic study of how market characteristics affect allocative performance and supporting evidence-based platform optimization in increasingly complex digital economies.

\section{Concluding remarks}

This paper presents a proof of concept. We have shown that it is possible to train predictive models for welfare and efficiency assessments using experimental data. 
We did so using data from experimental double auctions a la Vernon Smith (1962), and our prediction crucially does not rely on information regarding players' reservation values, as would typically be unobservable for real-world orderbook data. 
But why predict something that is actually known for this kind of data? We argue that the fact that our models are trained on this kind of controlled data makes these models interesting for future use on observational data, where the question we ask is not currently being addressed at all, because this question is critical yet very hard to answer based on conventional methods. The conceptual contribution to show a path to tackle such questions via experimental data is the main contribution of this article. 

Broadly summarizing the results, we found 
that non-linear methods performed well for predicting efficiency (AE) in general and also for predicting prices prior to any price realization based only on unfilled bid and ask data. These results and some preliminary results from running other predictive models (e.g., based on neural networks) suggest that more theory-informed/ model-based approaches may predict efficiency features more accurately, especially prior to any price realizations. We will continue to pursue developing such models, aiming to predict online in the future. Other next steps are (i.) to predict out-of-sample on other existing datasets for the same type of experimental double auction without performing training or hyper-parameter evaluation on them before, (ii.) to predict out-of-sample on data for the same kind of game that we shall collect going forward in a pre-registered prediction challenge, (iii.) then to predict out-of-sample on nonexperimental data but for the same kinds of markets as in the experiments (i.e. single-item continuous bid-ask double auctions for nondurables), which will require us to think about how such predictions on nonexperimental data can be validated. Beyond this, we also want to extend our scope to other types of markets, such as stock markets.

\section*{Declarations}

\subsection*{Availability of data and material}
This manuscript does not report newly generated data. The experimental dataset analysed is that of \citet{ikicacompeq}; the pre-processed data and the analysis code used to produce the results are publicly available at \url{https://github.com/asikist/inverse_experimental_markets}.

\subsection*{Competing interests}
The authors declare no competing interests as defined by the publisher, or any other interests that might be perceived to influence the results and/or discussion reported in this paper.

\subsection*{Funding}
Both authors acknowledge support from NCCR Automation, a National Centre of Competence in Research funded by the Swiss National Science Foundation (grant number 180545).

\subsection*{Authors' contributions}
T.A.\ contributed to the data analysis, preprocessing, and machine-learning modelling. H.N.\ provided the guidance, data, and economic perspective required for the inverse experimental analysis. Both authors contributed to drafting and revising the manuscript.

\subsection*{Acknowledgements}
We thank Prof.\ Rahul Savani for valuable input and feedback, and Dr.\ Barbara Ikica for data provision and support.

\clearpage

% Appendix
\appendix

\bibliographystyle{plainnat}
\bibliography{sn-bibliography}

\section*{Appendix}

\rev{The Python code and the pre-processed data are publicly available at \url{https://github.com/asikist/inverse_experimental_markets}.}

\section{Wilcoxon Tests for Model Comparison}
\bgroup
\def\arraystretch{1.2}%
\begin{table}[htb]
    \caption{\emph{Performance comparison for CE price prediction.} 
     Median APE CEP difference and paired Wilcoxon tests (significance - star) for different method pairs across time.
     The difference is calculated as the error of "Model A" minus the error "Model B" for the same test sample and the corresponding alternative hypothesis is tested based on the difference sign.
     Negative differences (bold) indicate that "Model A" in the column outperforms "Model B" in the row.
     }
     \label{tab:ce_ape:wilcoxon}
    \centering
    \small
        \begin{tabular}{@{\extracolsep{6pt}}cc | c | cccc}
        \toprule
         &  &  & \multicolumn{4}{c}{Median APE CEP Difference } \\ \hline
         & Price & Model A $\rightarrow$& EMH & CEMH & OB-RLM & GBT \\
        Round & realizations & Model B  $\downarrow$&  &  &  &  \\
        \midrule 
            \multirow[c]{8}{*}{1} & \multirow[c]{4}{*}{0} & EMH &   & n/a & \B -0.809\textsuperscript{*} & \B -0.865\textsuperscript{*} \\
             &  & CEMH & n/a &   & \B -0.809\textsuperscript{*} & \B -0.865\textsuperscript{*} \\
             &  & OB-RLM & 0.809\textsuperscript{*} & 0.809\textsuperscript{*} &   & \B -0.031\textsuperscript{*} \\
             &  & GBT & 0.865\textsuperscript{*} & 0.865\textsuperscript{*} & 0.031\textsuperscript{*} &   \\
            \cline{2-7}
             & \multirow[c]{4}{*}{1+} & EMH &   & \B -0.025\textsuperscript{*} & \B -0.047\textsuperscript{*} & \B -0.032\textsuperscript{*} \\
             &  & CEMH & 0.025\textsuperscript{*} &   & \B -0.026\textsuperscript{*} & \B -0.014\textsuperscript{*} \\
             &  & OB-RLM & 0.047\textsuperscript{*} & 0.026\textsuperscript{*} &   & 0.011\textsuperscript{*} \\
             &  & GBT & 0.032\textsuperscript{*} & 0.014\textsuperscript{*} & \B -0.011\textsuperscript{*} &   \\
            \cline{1-7} \cline{2-7}
            \multirow[c]{8}{*}{2+} & \multirow[c]{4}{*}{0} & EMH &   & n/a & \B -0.891\textsuperscript{*} & \B -0.901\textsuperscript{*} \\
             &  & CEMH & n/a &   & \B -0.891\textsuperscript{*} & \B -0.901\textsuperscript{*} \\
             &  & OB-RLM & 0.891\textsuperscript{*} & 0.891\textsuperscript{*} &   & \B -0.003\textsuperscript{*} \\
             &  & GBT & 0.901\textsuperscript{*} & 0.901\textsuperscript{*} & 0.003\textsuperscript{*} &   \\
            \cline{2-7}
             & \multirow[c]{4}{*}{1+} & EMH &   & \B -0.010\textsuperscript{*} & \B -0.007\textsuperscript{*} & \B -0.006\textsuperscript{*} \\
             &  & CEMH & 0.010\textsuperscript{*} &   & \B -0.005\textsuperscript{*} & \B -0.004\textsuperscript{*} \\
             &  & OB-RLM & 0.007\textsuperscript{*} & 0.005\textsuperscript{*} &   & 0.002\textsuperscript{*} \\
             &  & GBT & 0.006\textsuperscript{*} & 0.004\textsuperscript{*} & \B -0.002\textsuperscript{*} &   \\
        \bottomrule
        \end{tabular}
\end{table}
\egroup

\bgroup
\def\arraystretch{1.2}%
\begin{table}[htb]
    \caption{\emph{Performance comparison for AE prediction.} 
             Median APE CEP difference and paired Wilcoxon tests (significance stars) for different method pairs across time.
             The difference is calculated as the error of "Model A" minus the error "Model B" for the same test sample and the corresponding alternative hypothesis is tested based on the difference sign.
             Negative differences (bold) indicate that "Model A" in the column outperforms "Model B" in the row.
             }
    \label{tab:ae_ape:wilcoxon}
    \centering
    \small
        \begin{tabular}{@{\extracolsep{6pt}}cc | c | cccc}
        \toprule
         &  &  & \multicolumn{4}{c}{Median APE CEP Difference } \\ \hline
         & Price & Model A $\rightarrow$& EMH & CEMH & OB-RLM & GBT \\
        Round & realizations & Model B  $\downarrow$&  &  &  &  \\
        \midrule  
        \multirow[c]{8}{*}{1} & \multirow[c]{4}{*}{0} & EMH & - & \B -0.092\textsuperscript{*} & \B -0.113\textsuperscript{*} & \B -0.124\textsuperscript{*} \\
         &  & CEMH & 0.092\textsuperscript{*} & - & 0.023\textsuperscript{*} & \B -0.001\textsuperscript{*} \\
         &  & OB-RLM & 0.113\textsuperscript{*} & \B -0.023\textsuperscript{*} & - & \B -0.029\textsuperscript{*} \\
         &  & GBT & 0.124\textsuperscript{*} & 0.001\textsuperscript{*} & 0.029\textsuperscript{*} & - \\
        \cline{2-7}
         & \multirow[c]{4}{*}{1+} & EMH & - & \B -0.114\textsuperscript{*} & \B -0.125\textsuperscript{*} & \B -0.148\textsuperscript{*} \\
         &  & CEMH & 0.114\textsuperscript{*} & - & 0.044\textsuperscript{*} & \B -0.003\textsuperscript{*} \\
         &  & OB-RLM & 0.125\textsuperscript{*} & \B -0.044\textsuperscript{*} & - & \B -0.046\textsuperscript{*} \\
         &  & GBT & 0.148\textsuperscript{*} & 0.003\textsuperscript{*} & 0.046\textsuperscript{*} & - \\
        \cline{1-7} \cline{2-7}
        \multirow[c]{8}{*}{2+} & \multirow[c]{4}{*}{0} & EMH & - & \B -0.025\textsuperscript{*} & \B -0.034\textsuperscript{*} & \B -0.055\textsuperscript{*} \\
         &  & CEMH & 0.025\textsuperscript{*} & - & 0.039\textsuperscript{*} & 0.000 \\
         &  & OB-RLM & 0.034\textsuperscript{*} & \B -0.039\textsuperscript{*} & - & \B -0.034\textsuperscript{*} \\
         &  & GBT & 0.055\textsuperscript{*} & 0.000 & 0.034\textsuperscript{*} & - \\
        \cline{2-7}
         & \multirow[c]{4}{*}{1+} & EMH & - & \B -0.027\textsuperscript{*} & \B -0.034\textsuperscript{*} & \B -0.051\textsuperscript{*} \\
         &  & CEMH & 0.027\textsuperscript{*} & - & 0.021\textsuperscript{*} & \B -0.002\textsuperscript{*} \\
         &  & OB-RLM & 0.034\textsuperscript{*} & \B -0.021\textsuperscript{*} & - & \B -0.023\textsuperscript{*} \\
         &  & GBT & 0.051\textsuperscript{*} & 0.002\textsuperscript{*} & 0.023\textsuperscript{*} & - \\
            \bottomrule
        \end{tabular}
\end{table}
\egroup

\clearpage
\section{\rev{Session-Clustered Wilcoxon Tests}}\label{sec:appendix:wilcoxon:clustered}

\rev{The Wilcoxon tests of \cref{tab:ce_ape:wilcoxon,tab:ae_ape:wilcoxon} treat every $(\mathit{sample\_id}, \mathit{treatment}, \mathit{game}, \mathit{round}, \mathit{time})$ row as an independent observation. Rows from the same experimental game share valuations, players and a common realised orderbook, so the independence assumption is violated and the reported p-values are anti-conservative. We therefore report two additional variants that treat the game as the unit of analysis. In both, the cluster is $(\mathit{treatment}, \mathit{game})$ with $K$ clusters ranging from $86$ to $89$ across buckets, and the alternative is two-sided. P-values are Holm-adjusted within each task across the $24$ comparisons ($4$ buckets $\times$ $6$ model pairs) so that each cited cell controls family-wise error at the $5\%$ level.}

\rev{\emph{Clustered signed-rank (RGL).} For each $(\text{Round}, \text{Deals})$ bucket and each model pair we form paired differences $d = \mathrm{APE}_A - \mathrm{APE}_B$ for all rows in the bucket, then run the Rosner-Glynn-Lee clustered signed-rank test~\cite{rosner1999use} as implemented in the R package \texttt{clusrank}~\cite{jiang2020clusrank}. The test statistic sums signed ranks within clusters, and its variance is estimated from the cluster-level sums, so it retains within-cluster information while accounting for within-cluster correlation. Results are reported in \cref{tab:cep:wilcoxon:clustered,tab:ae:wilcoxon:clustered}.}

\rev{\emph{Median-aggregated.} We collapse to one value per game by taking the per-game median of $d$ within the bucket and run a standard two-sided paired Wilcoxon signed-rank test (\texttt{scipy.stats.wilcoxon}) across the $\sim 88$ game-level medians. This discards within-cluster information but makes independence exact at the game level. Results are reported in \cref{tab:cep:wilcoxon:aggregated,tab:ae:wilcoxon:aggregated}.}

\rev{\emph{Note on the reported medians.} The cell values in \cref{tab:cep:wilcoxon:clustered,tab:ae:wilcoxon:clustered,tab:cep:wilcoxon:aggregated,tab:ae:wilcoxon:aggregated} are identical to those in \cref{tab:ce_ape:wilcoxon,tab:ae_ape:wilcoxon} and coincide across the clustered and median-aggregated variants because we always report the same bucket-level median of $\mathrm{APE}_A - \mathrm{APE}_B$ as a point summary; only the significance stars differ, and they reflect the different p-values produced by each test.}

\rev{The two variants broadly agree: after Holm adjustment, the clustered test flags $16/24$ (AE) and $14/24$ (CEP) comparisons as significant at $5\%$; the median-aggregated test flags $15/24$ and $14/24$. The substantive ranking of \cref{tab:ae_ape,tab:cep_ape} is preserved in both. The comparisons that lose significance under clustering are exactly those where the original per-row test was driven by very small effect sizes amplified by pseudo-replication (e.g. CEMH vs.\ GBT cells, where the median difference rounds to $\pm 0.001$--$0.003$).}

\begin{table}[htb]
\centering
\caption{\emph{\rev{Clustered signed-rank test, CEP.}} \rev{Median $(\mathrm{APE}_A - \mathrm{APE}_B)$ per bucket, with RGL clustered signed-rank test on all rows in the bucket (cluster $= (\mathit{treatment}, \mathit{game})$, two-sided, Holm-adjusted across $24$ comparisons). Significance stars on the Holm-adjusted $p$: $*\,p<0.1$; $**\,p<0.05$; $***\,p<0.01$. Negative cells (bold) indicate Model A (column) beats Model B (row).}}
\label{tab:cep:wilcoxon:clustered}
\small
\begin{tabular}{lllllll}
\toprule
 &  &  & \multicolumn{4}{r}{Median APE Difference} \\
 &  & Model A & EMH & CEMH & OB-RLM & GBT \\
Round & Price realizations & Model B &  &  &  &  \\
\midrule
\multirow[c]{8}{*}{1} & \multirow[c]{4}{*}{0} & EMH &  & n/a & \B -0.809\textsuperscript{***} & \B -0.865\textsuperscript{***} \\
 &  & CEMH & n/a &  & \B -0.809\textsuperscript{***} & \B -0.865\textsuperscript{***} \\
 &  & OB-RLM & 0.809\textsuperscript{***} & 0.809\textsuperscript{***} &  & \B -0.031\textsuperscript{***} \\
 &  & GBT & 0.865\textsuperscript{***} & 0.865\textsuperscript{***} & 0.031\textsuperscript{***} &  \\
\cline{2-7}
 & \multirow[c]{4}{*}{1+} & EMH &  & \B -0.025\textsuperscript{***} & \B -0.047\textsuperscript{***} & \B -0.032\textsuperscript{***} \\
 &  & CEMH & 0.025\textsuperscript{***} &  & \B -0.026\textsuperscript{***} & \B -0.014\textsuperscript{**} \\
 &  & OB-RLM & 0.047\textsuperscript{***} & 0.026\textsuperscript{***} &  & 0.011 \\
 &  & GBT & 0.032\textsuperscript{***} & 0.014\textsuperscript{**} & \B -0.011 &  \\
\cline{1-7} \cline{2-7}
\multirow[c]{8}{*}{2+} & \multirow[c]{4}{*}{0} & EMH &  & n/a & \B -0.891\textsuperscript{***} & \B -0.901\textsuperscript{***} \\
 &  & CEMH & n/a &  & \B -0.891\textsuperscript{***} & \B -0.901\textsuperscript{***} \\
 &  & OB-RLM & 0.891\textsuperscript{***} & 0.891\textsuperscript{***} &  & \B -0.003 \\
 &  & GBT & 0.901\textsuperscript{***} & 0.901\textsuperscript{***} & 0.003 &  \\
\cline{2-7}
 & \multirow[c]{4}{*}{1+} & EMH &  & \B -0.010 & \B -0.007 & \B -0.006 \\
 &  & CEMH & 0.010 &  & \B -0.005 & \B -0.004 \\
 &  & OB-RLM & 0.007 & 0.005 &  & 0.002 \\
 &  & GBT & 0.006 & 0.004 & \B -0.002 &  \\
\cline{1-7} \cline{2-7}
\bottomrule
\end{tabular}

\end{table}

\begin{table}[htb]
\centering
\caption{\emph{\rev{Clustered signed-rank test, AE.}} \rev{As in \cref{tab:cep:wilcoxon:clustered} but for the AE task.}}
\label{tab:ae:wilcoxon:clustered}
\small
\begin{tabular}{lllllll}
\toprule
 &  &  & \multicolumn{4}{r}{Median APE Difference} \\
 &  & Model A & EMH & CEMH & OB-RLM & GBT \\
Round & \# Total Deals & Model B &  &  &  &  \\
\midrule
\multirow[c]{8}{*}{1} & \multirow[c]{4}{*}{0} & EMH &  & \B -0.092\textsuperscript{***} & \B -0.113\textsuperscript{**} & \B -0.124\textsuperscript{***} \\
 &  & CEMH & 0.092\textsuperscript{***} &  & 0.023\textsuperscript{**} & \B -0.001 \\
 &  & OB-RLM & 0.113\textsuperscript{**} & \B -0.023\textsuperscript{**} &  & \B -0.029\textsuperscript{***} \\
 &  & GBT & 0.124\textsuperscript{***} & 0.001 & 0.029\textsuperscript{***} &  \\
\cline{2-7}
 & \multirow[c]{4}{*}{1+} & EMH &  & \B -0.114\textsuperscript{***} & \B -0.125\textsuperscript{***} & \B -0.148\textsuperscript{***} \\
 &  & CEMH & 0.114\textsuperscript{***} &  & 0.044 & \B -0.003 \\
 &  & OB-RLM & 0.125\textsuperscript{***} & \B -0.044 &  & \B -0.046\textsuperscript{**} \\
 &  & GBT & 0.148\textsuperscript{***} & 0.003 & 0.046\textsuperscript{**} &  \\
\cline{1-7} \cline{2-7}
\multirow[c]{8}{*}{2+} & \multirow[c]{4}{*}{0} & EMH &  & \B -0.025\textsuperscript{***} & \B -0.034 & \B -0.055\textsuperscript{***} \\
 &  & CEMH & 0.025\textsuperscript{***} &  & 0.039\textsuperscript{***} & n/a \\
 &  & OB-RLM & 0.034 & \B -0.039\textsuperscript{***} &  & \B -0.034\textsuperscript{***} \\
 &  & GBT & 0.055\textsuperscript{***} & n/a & 0.034\textsuperscript{***} &  \\
\cline{2-7}
 & \multirow[c]{4}{*}{1+} & EMH &  & \B -0.027\textsuperscript{***} & \B -0.034\textsuperscript{**} & \B -0.051\textsuperscript{***} \\
 &  & CEMH & 0.027\textsuperscript{***} &  & 0.021\textsuperscript{*} & \B -0.002 \\
 &  & OB-RLM & 0.034\textsuperscript{**} & \B -0.021\textsuperscript{*} &  & \B -0.023 \\
 &  & GBT & 0.051\textsuperscript{***} & 0.002 & 0.023 &  \\
\cline{1-7} \cline{2-7}
\bottomrule
\end{tabular}

\end{table}

\begin{table}[htb]
\centering
\caption{\emph{\rev{Median-aggregated signed-rank test, CEP.}} \rev{Same bucketing as \cref{tab:cep:wilcoxon:clustered}; test is two-sided paired Wilcoxon on the $\sim 88$ per-game median differences, Holm-adjusted. Bold negative = Model A beats Model B. The test discards within-game information; power is lower than the RGL variant but independence is exact at the game level.}}
\label{tab:cep:wilcoxon:aggregated}
\small
\begin{tabular}{lllllll}
\toprule
 &  &  & \multicolumn{4}{r}{Median APE Difference} \\
 &  & Model A & EMH & CEMH & OB-RLM & GBT \\
Round & Price realizations & Model B &  &  &  &  \\
\midrule
\multirow[c]{8}{*}{1} & \multirow[c]{4}{*}{0} & EMH &  & n/a & \B -0.809\textsuperscript{***} & \B -0.865\textsuperscript{***} \\
 &  & CEMH & n/a &  & \B -0.809\textsuperscript{***} & \B -0.865\textsuperscript{***} \\
 &  & OB-RLM & 0.809\textsuperscript{***} & 0.809\textsuperscript{***} &  & \B -0.031\textsuperscript{**} \\
 &  & GBT & 0.865\textsuperscript{***} & 0.865\textsuperscript{***} & 0.031\textsuperscript{**} &  \\
\cline{2-7}
 & \multirow[c]{4}{*}{1+} & EMH &  & \B -0.025\textsuperscript{***} & \B -0.047\textsuperscript{***} & \B -0.032\textsuperscript{***} \\
 &  & CEMH & 0.025\textsuperscript{***} &  & \B -0.026\textsuperscript{***} & \B -0.014\textsuperscript{***} \\
 &  & OB-RLM & 0.047\textsuperscript{***} & 0.026\textsuperscript{***} &  & 0.011 \\
 &  & GBT & 0.032\textsuperscript{***} & 0.014\textsuperscript{***} & \B -0.011 &  \\
\cline{1-7} \cline{2-7}
\multirow[c]{8}{*}{2+} & \multirow[c]{4}{*}{0} & EMH &  & n/a & \B -0.891\textsuperscript{***} & \B -0.901\textsuperscript{***} \\
 &  & CEMH & n/a &  & \B -0.891\textsuperscript{***} & \B -0.901\textsuperscript{***} \\
 &  & OB-RLM & 0.891\textsuperscript{***} & 0.891\textsuperscript{***} &  & \B -0.003 \\
 &  & GBT & 0.901\textsuperscript{***} & 0.901\textsuperscript{***} & 0.003 &  \\
\cline{2-7}
 & \multirow[c]{4}{*}{1+} & EMH &  & \B -0.010 & \B -0.007 & \B -0.006 \\
 &  & CEMH & 0.010 &  & \B -0.005\textsuperscript{*} & \B -0.004 \\
 &  & OB-RLM & 0.007 & 0.005\textsuperscript{*} &  & 0.002 \\
 &  & GBT & 0.006 & 0.004 & \B -0.002 &  \\
\cline{1-7} \cline{2-7}
\bottomrule
\end{tabular}

\end{table}

\begin{table}[htb]
\centering
\caption{\emph{\rev{Median-aggregated signed-rank test, AE.}} \rev{As in \cref{tab:cep:wilcoxon:aggregated} but for the AE task.}}
\label{tab:ae:wilcoxon:aggregated}
\small
\begin{tabular}{lllllll}
\toprule
 &  &  & \multicolumn{4}{r}{Median APE Difference} \\
 &  & Model A & EMH & CEMH & OB-RLM & GBT \\
Round & \# Total Deals & Model B &  &  &  &  \\
\midrule
\multirow[c]{8}{*}{1} & \multirow[c]{4}{*}{0} & EMH &  & \B -0.092\textsuperscript{***} & \B -0.113\textsuperscript{***} & \B -0.124\textsuperscript{***} \\
 &  & CEMH & 0.092\textsuperscript{***} &  & 0.023 & \B -0.001 \\
 &  & OB-RLM & 0.113\textsuperscript{***} & \B -0.023 &  & \B -0.029\textsuperscript{**} \\
 &  & GBT & 0.124\textsuperscript{***} & 0.001 & 0.029\textsuperscript{**} &  \\
\cline{2-7}
 & \multirow[c]{4}{*}{1+} & EMH &  & \B -0.114\textsuperscript{***} & \B -0.125\textsuperscript{***} & \B -0.148\textsuperscript{***} \\
 &  & CEMH & 0.114\textsuperscript{***} &  & 0.044 & \B -0.003 \\
 &  & OB-RLM & 0.125\textsuperscript{***} & \B -0.044 &  & \B -0.046\textsuperscript{**} \\
 &  & GBT & 0.148\textsuperscript{***} & 0.003 & 0.046\textsuperscript{**} &  \\
\cline{1-7} \cline{2-7}
\multirow[c]{8}{*}{2+} & \multirow[c]{4}{*}{0} & EMH &  & \B -0.025\textsuperscript{***} & \B -0.034 & \B -0.055\textsuperscript{***} \\
 &  & CEMH & 0.025\textsuperscript{***} &  & 0.039\textsuperscript{**} & n/a \\
 &  & OB-RLM & 0.034 & \B -0.039\textsuperscript{**} &  & \B -0.034\textsuperscript{***} \\
 &  & GBT & 0.055\textsuperscript{***} & n/a & 0.034\textsuperscript{***} &  \\
\cline{2-7}
 & \multirow[c]{4}{*}{1+} & EMH &  & \B -0.027\textsuperscript{***} & \B -0.034\textsuperscript{**} & \B -0.051\textsuperscript{***} \\
 &  & CEMH & 0.027\textsuperscript{***} &  & 0.021\textsuperscript{*} & \B -0.002 \\
 &  & OB-RLM & 0.034\textsuperscript{**} & \B -0.021\textsuperscript{*} &  & \B -0.023 \\
 &  & GBT & 0.051\textsuperscript{***} & 0.002 & 0.023 &  \\
\cline{1-7} \cline{2-7}
\bottomrule
\end{tabular}

\end{table}

\clearpage
\section{\rev{Voronoi spatial partitioning}}\label{sec:appendix:voronoi}

\rev{This appendix collects the spatial breakdown of the per-game best-performing model across buyer and seller valuation-decile distances summarised in \cref{sec:results:analysis,sec:cep:pred:analysis}. Each scatter point is one (game, round) pair; coordinates are the sum of decile differences between the realised buyer (resp.\ seller) valuation distribution and the round median. Points are then enclosed in Voronoi cells to highlight contiguous regions in which a single model type is dominant. When several markets coincide on the decile grid we add a small Gaussian jitter to make the partition well defined.}

\subsection{\rev{Allocative efficiency}}\label{sec:appendix:voronoi:ae}

\rev{Non-linear models occupy the upper-left quadrant in the first round before any price realisation, which is also the densest region of the dataset (markets with lower-than-median bid and ask prices). As rounds advance and realised prices accumulate, the linear models reclaim share, consistent with player behaviour shifting the relationship between observed orderbook data and AE.}

\begin{figure}[htb!]
    \subfloat[First round, prior to any price realization]{
        \includegraphics[width=0.475\linewidth]{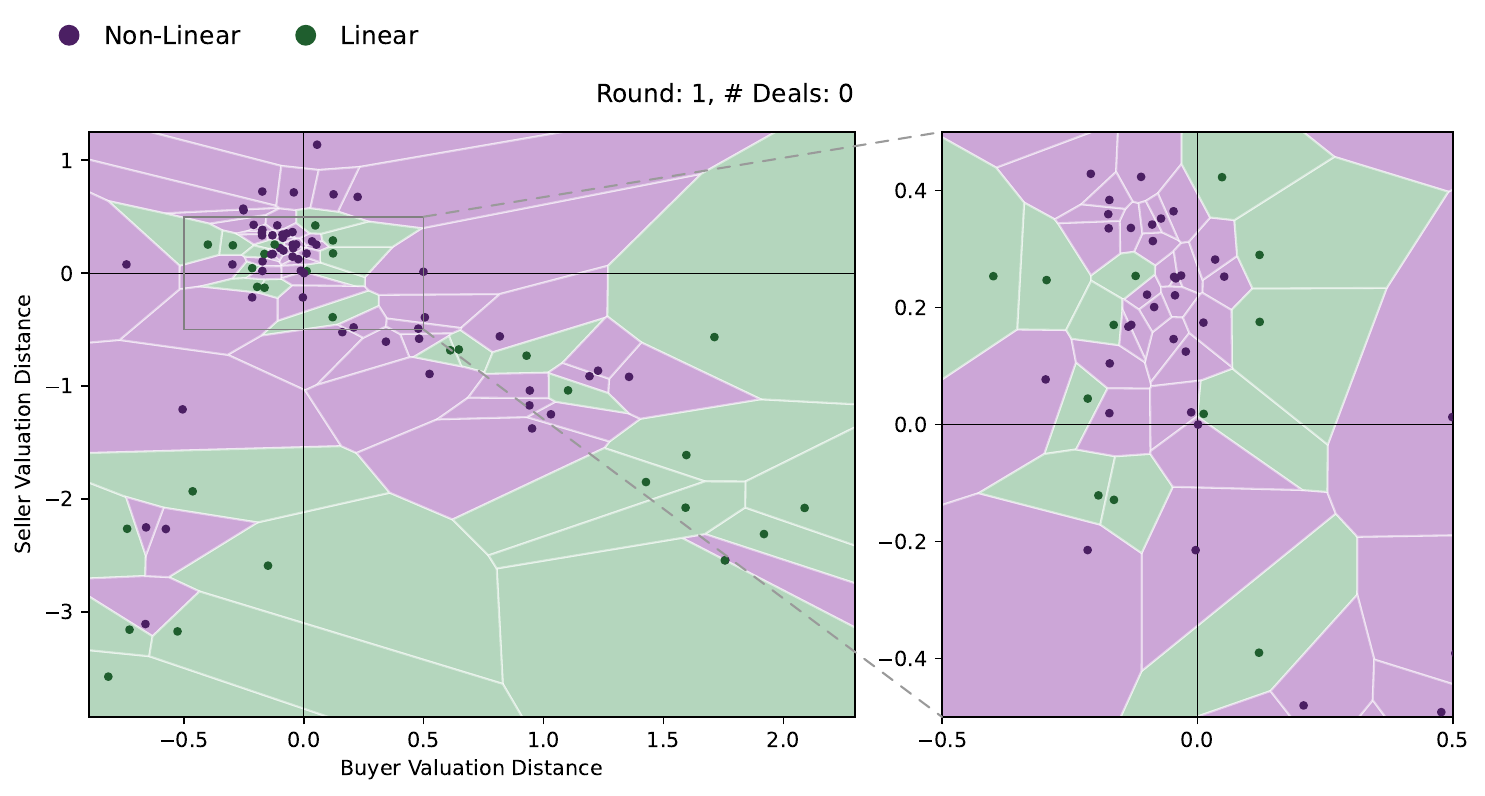}
        \label{fig:ae:distance:voronoi:1}
    }
    \subfloat[First rounds, at least one price realization]{
        \includegraphics[width=0.475\linewidth]{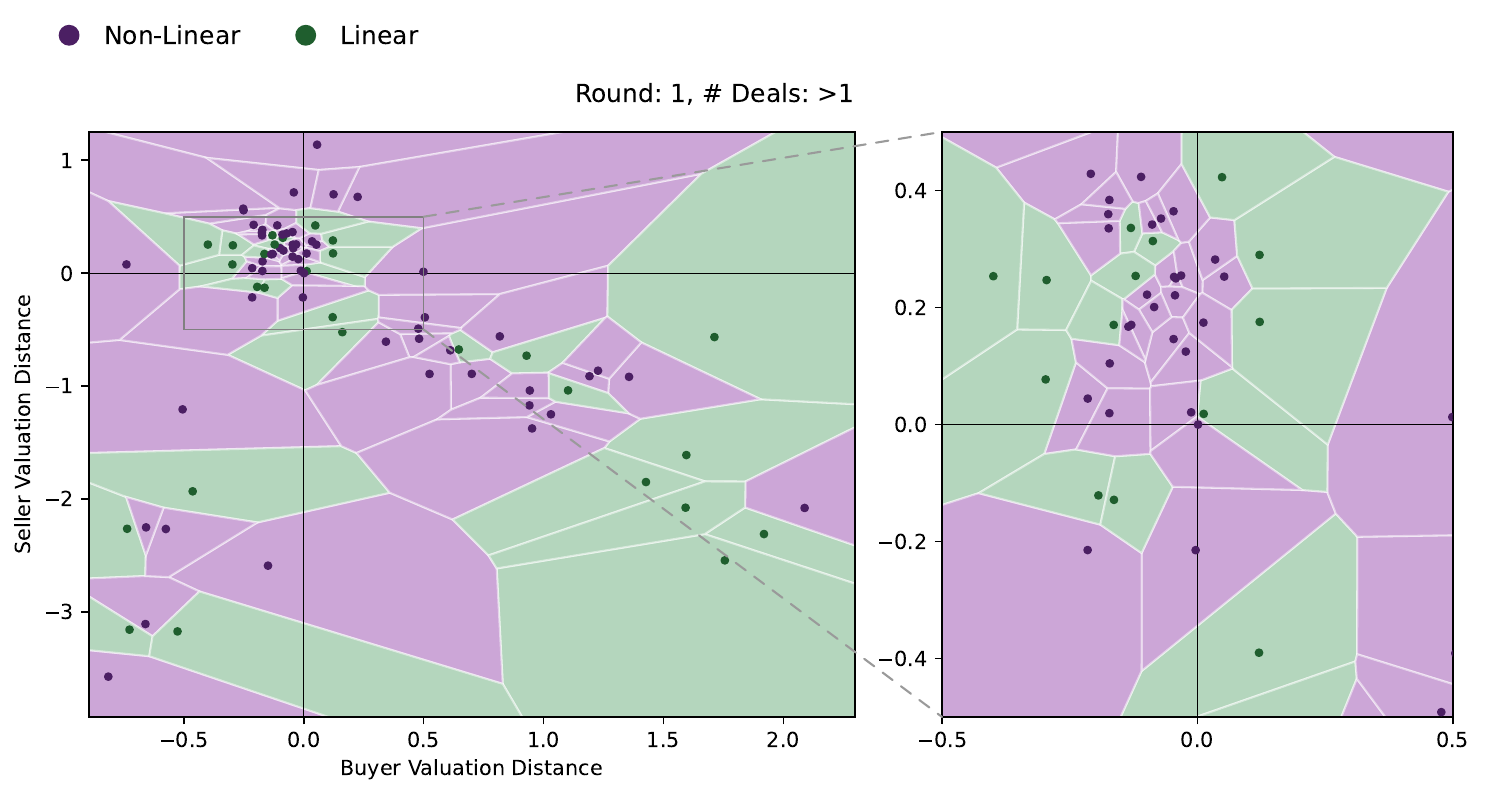}
        \label{fig:ae:distance:voronoi:2}
    }\\
    \subfloat[2+ rounds, prior to any price realization]{
        \includegraphics[width=0.475\linewidth]{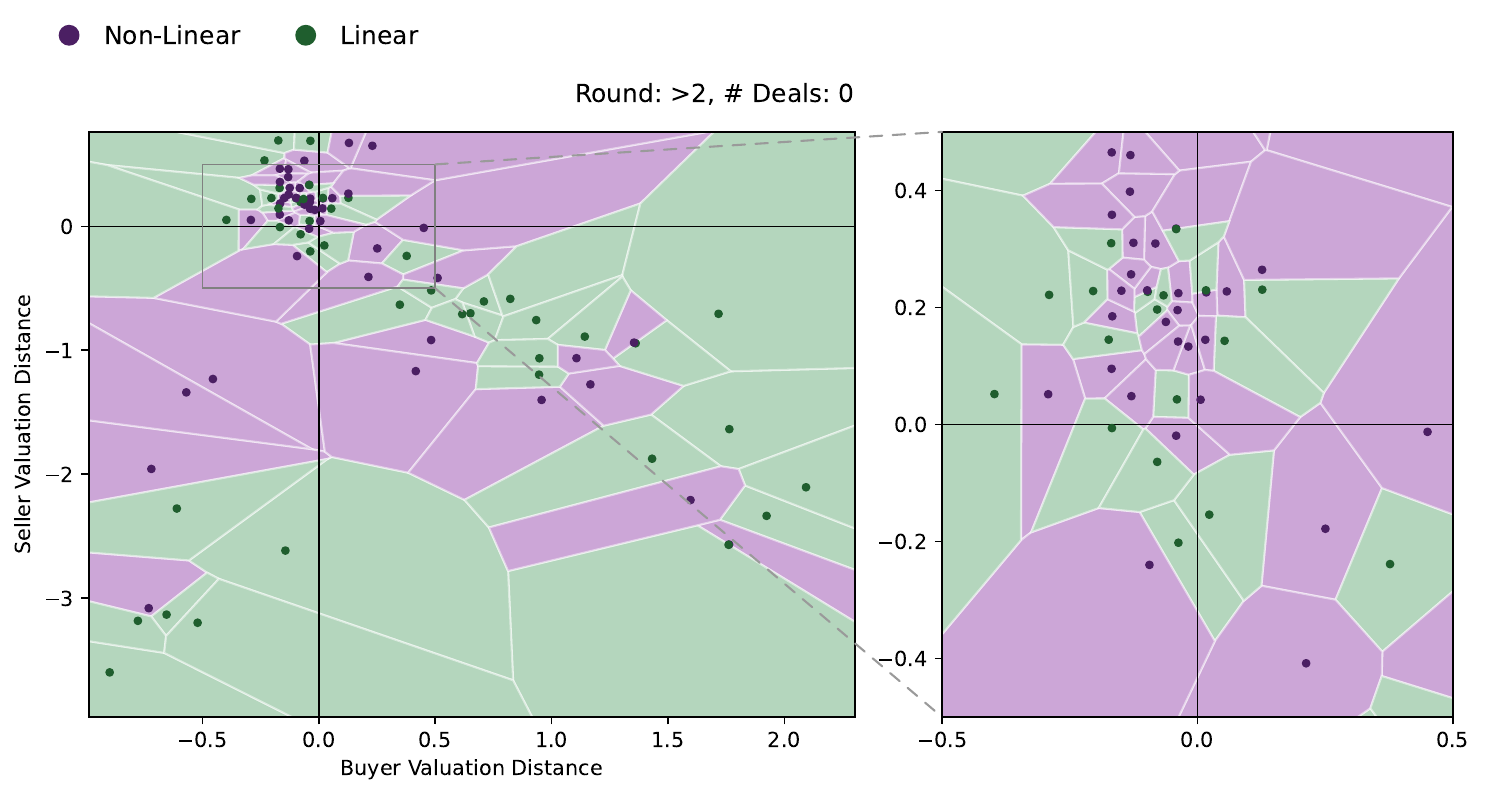}
        \label{fig:ae:distance:voronoi:3}
    }
    \subfloat[2+ rounds, at least one price realization]{
        \includegraphics[width=0.475\linewidth]{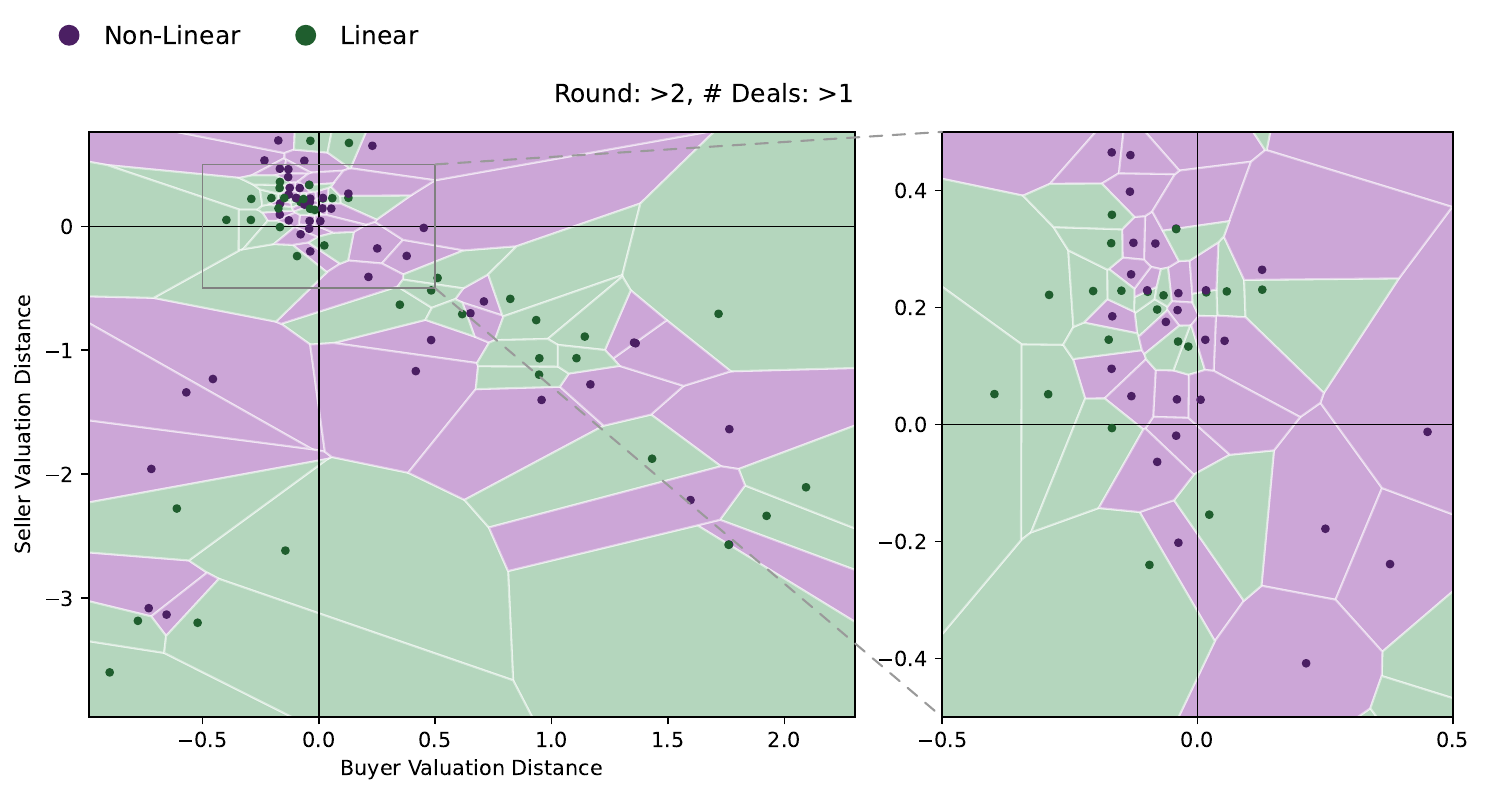}
        \label{fig:ae:distance:voronoi:4}
    }\\
    \caption{\emph{Top allocative efficiency performing models, based on model type, before and after initial price realizations, as a function of the sum of differences (market distance in terms) of buyer/seller valuation deciles.} Comparison of top ranking models in terms of MAPE values for AE for markets that have different valuation price profiles. Each scatter point is included in a voronoi area, in order to detect continuous areas of similar behavior. When two different markets have the same distance coordinates in both axis, we differentiate them for the vornoi by adding a small Gaussian noise to their coordinates.}
    \label{fig:ae:distance:voronoi}
\end{figure}

\subsection{\rev{Competitive equilibrium price}}\label{sec:appendix:voronoi:cep}

\rev{For CEP, GBT (the only non-linear model) dominates contiguous regions of the decile-distance plane before any deal is realised. Once realised prices accumulate within a round, the OB-RLM linear model reclaims the same regions, mirroring the AE pattern.}

\begin{figure}[htb!]
    \subfloat[First round, prior to any price realization]{
        \includegraphics[width=0.475\linewidth]{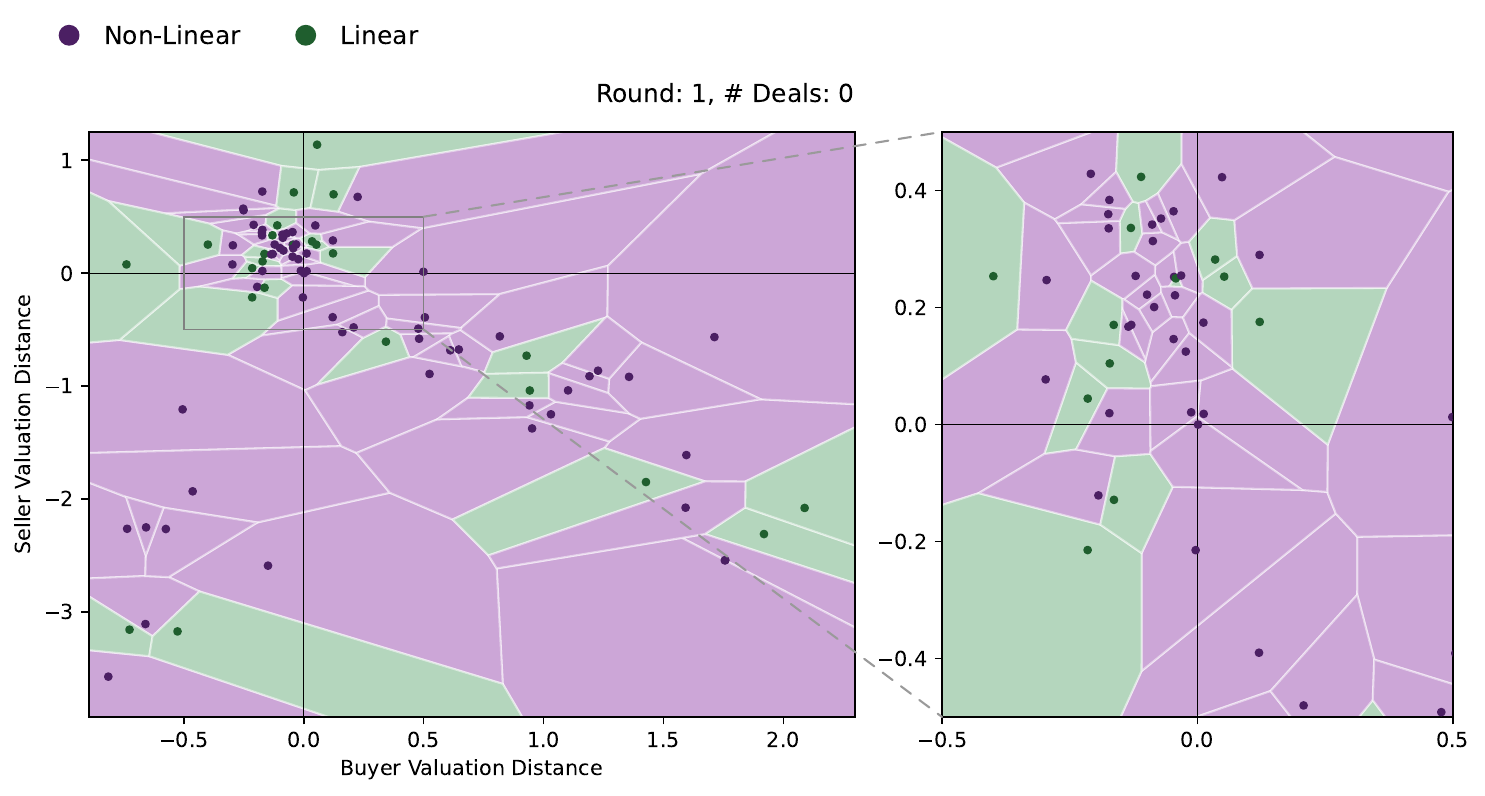}
        \label{fig:cep:distance:voronoi:1}
    }
    \subfloat[First rounds, at least one price realization]{
        \includegraphics[width=0.475\linewidth]{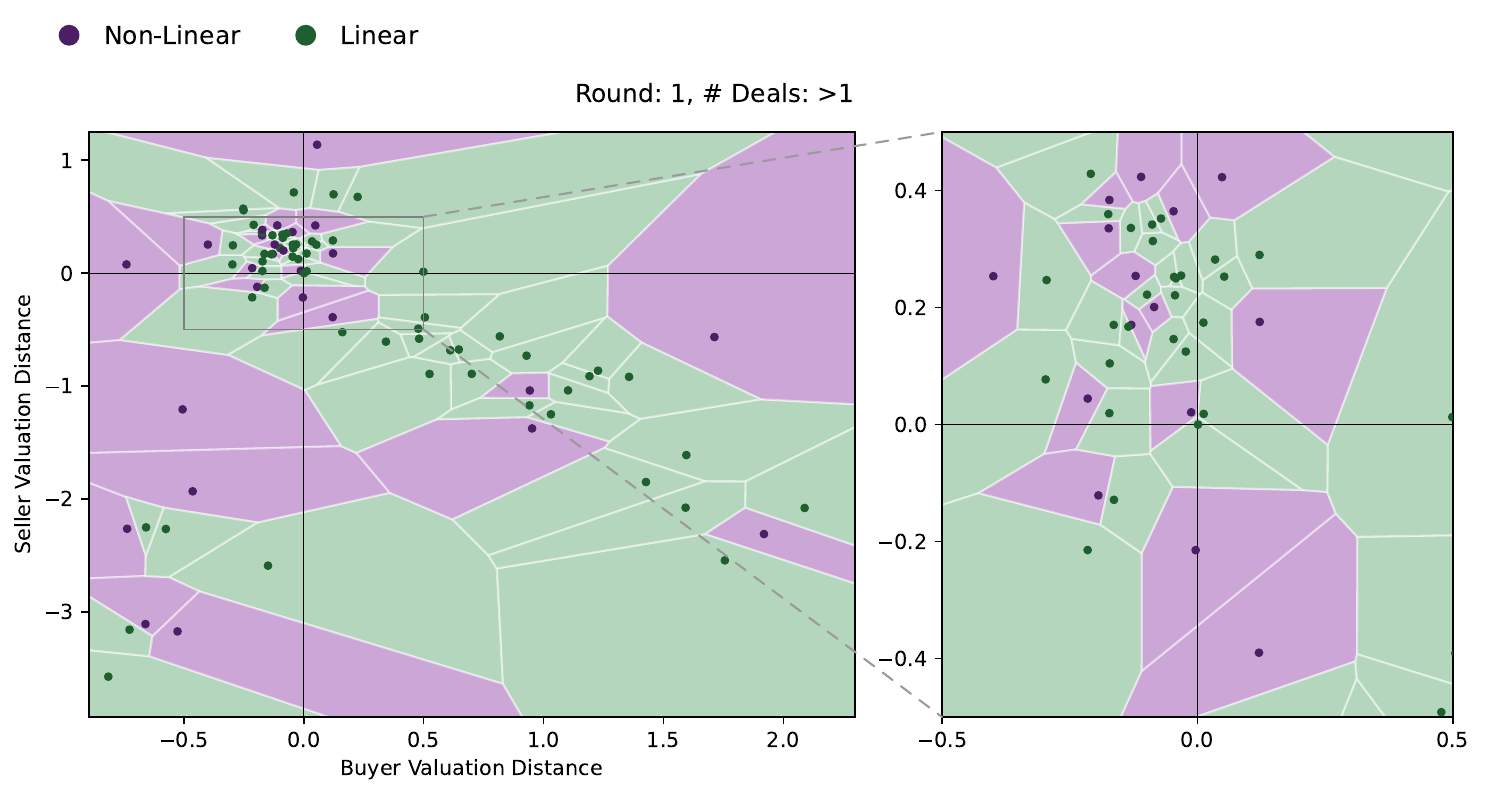}
        \label{fig:cep:distance:voronoi:2}
    }\\
    \subfloat[2+ rounds, prior to any price realization]{
        \includegraphics[width=0.475\linewidth]{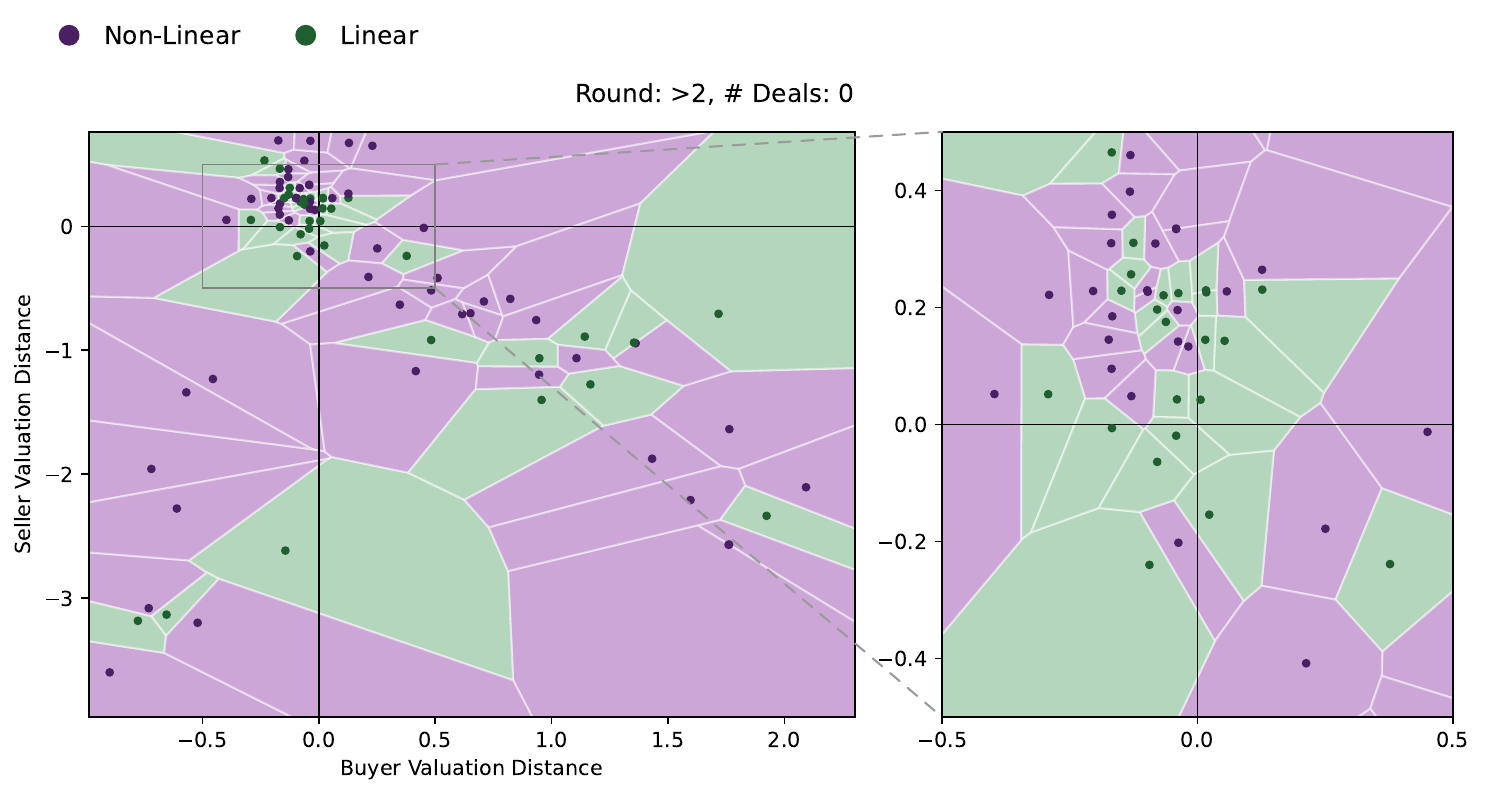}
        \label{fig:cep:distance:voronoi:3}
    }
    \subfloat[2+ rounds, at least one price realization]{
        \includegraphics[width=0.475\linewidth]{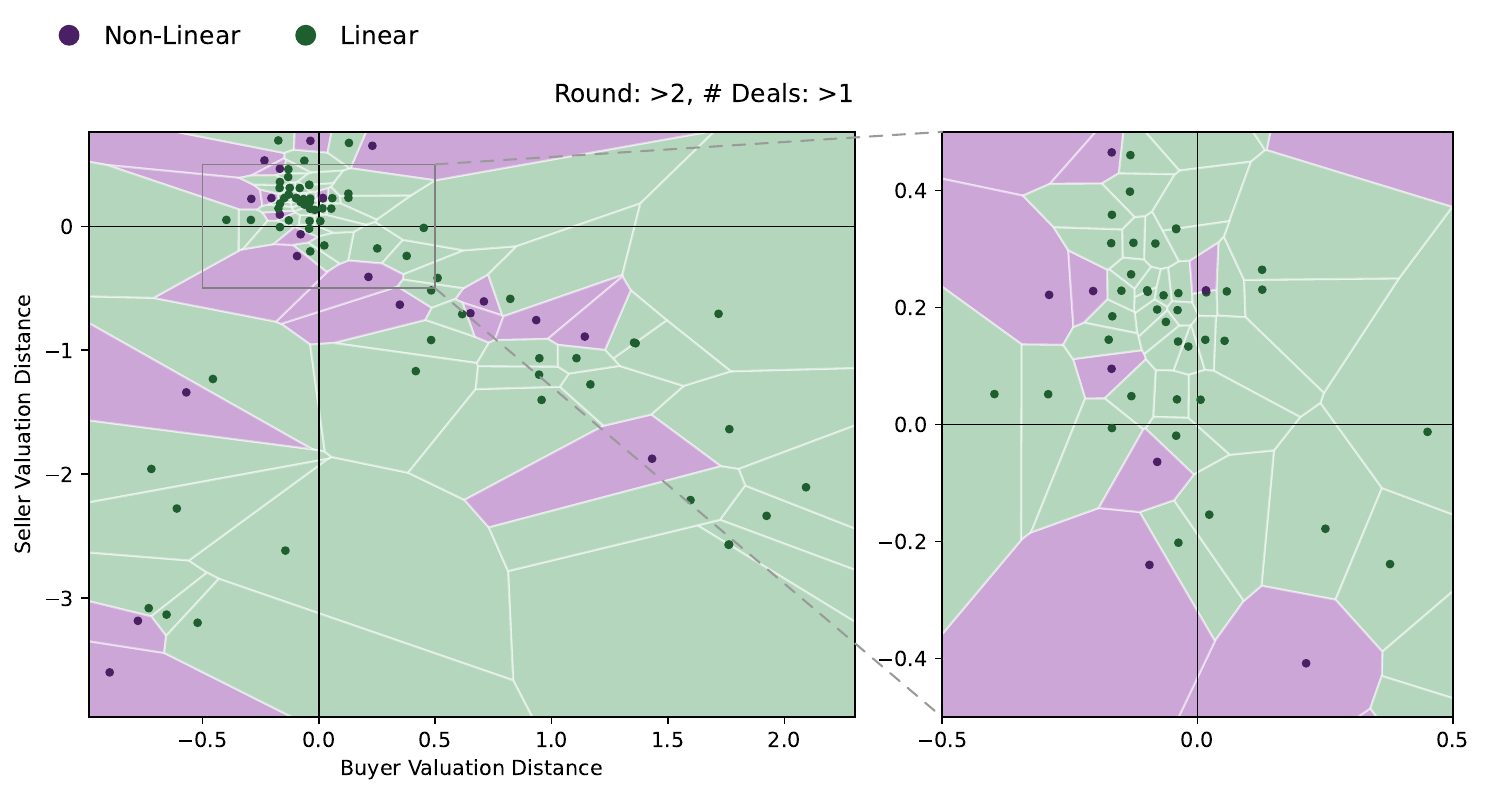}
        \label{fig:cep:distance:voronoi:4}
    }\\
    \caption{\emph{Top CEP performing models, based on model type, before and after initial price realizations, as a function of market distance in terms of valuation deciles.} Comparison of top ranking models in terms of MAPE values for CEP for markets that have different valuation price profiles. Each scatter point is included in a voronoi area, in order to detect continuous areas of dominant areas. For games falling on same distance coordinates, due to having same valuation deciles, we use a small amount of Gaussian jittering. In this plot, only the GBT is treated as non-linear model.}
    \label{fig:cep:distance:voronoi}
\end{figure}

\clearpage
\section{\rev{Simple CEP Baselines}}
\label{sec:appendix:cep:baselines}

To contextualise the CEP results of \cref{tab:cep_ape}, we also evaluate two further simple baselines.
The first is a \emph{persistence} predictor, which predicts the last observed realised price.
This is exactly the EMH model as defined in \cref{sec:cemh:price}, so it is already included in the main comparison and we do not duplicate it here.
The other two predictors we consider are:

\begin{itemize}
  \item \textbf{Treatment-Mean}, a constant prediction per treatment equal to the mean of $\mathit{ce\_round}$ computed over distinct (game, round) pairs in the training split. It uses only the treatment label at prediction time.
  \item \textbf{Book-Midpoint}, the average $(\text{best bid} + \text{best ask})/2$ at prediction time, where the best bid is the highest buyer-bid decile $\vquantile{\vbid}{1.0}$ and the best ask is the lowest seller-ask decile $\vquantile{\vask}{0.0}$. When one side of the book is empty we use the other side; when both sides are empty we fall back to the Treatment-Mean prediction.
\end{itemize}

Both predictors are implemented in \texttt{code/scripts/cep\_baselines.py} and evaluated on the same per-sample-id test rows as the four models retained in the main text.
\cref{tab:cep:baseline:comparison} reports their median CEP APE on the same four (Round, Price realizations) buckets used in \cref{tab:cep_ape}.

\begin{table}[htb]
\centering
\caption{\emph{Median APE of the four paper models and the two further predictors on the same test rows.} Buckets are the same as in \cref{tab:cep_ape}. Italicised rows are the two predictors considered only in this appendix.}
\label{tab:cep:baseline:comparison}
\small
\begin{tabular}{l cccc}
\toprule
 & \multicolumn{2}{c}{Round 1} & \multicolumn{2}{c}{Round $\geq 2$} \\
\cmidrule(lr){2-3}\cmidrule(lr){4-5}
Model & 0 deals & $\geq 1$ deal & 0 deals & $\geq 1$ deal \\
\midrule
EMH & 1.000 & 0.109 & 1.000 & 0.062 \\
CEMH & 1.000 & 0.092 & 1.000 & 0.055 \\
OB-RLM & 0.191 & 0.061 & 0.109 & 0.048 \\
GBT & 0.135 & 0.077 & 0.099 & 0.051 \\
\textit{Treatment-Mean} & 0.054 & 0.050 & 0.050 & 0.050 \\
\textit{Book-Midpoint} & 0.286 & 0.100 & 0.103 & 0.059 \\
\bottomrule
\end{tabular}

\end{table}

We observe two patterns in \cref{tab:cep:baseline:comparison}.
First, Book-Midpoint is uniformly dominated by OB-RLM and GBT, and is also dominated by Treatment-Mean in every bucket. It does not alter the conclusions of the main text.
Second, Treatment-Mean reports a nearly flat median APE of $\sim 0.05$ across all four buckets, and in the no-deals columns it is numerically smaller than EMH, CEMH, OB-RLM and GBT.
At first reading this appears to contradict the main results of \cref{tab:cep_ape}; two observations resolve the apparent tension.

Since $\mathit{ce\_round}$ is narrowly dispersed within a treatment by construction (each treatment fixes a reservation-value schedule, and per-round CE variation comes only from which subset of buyers and sellers is active in that round), a constant prediction equal to the within-treatment mean achieves a median APE close to the coefficient of variation of $\mathit{ce\_round}$ within a treatment, which is around $5\%$.
The orderbook-based models do substantially more than reproducing this prior. They track within-game dispersion, tails, and regime shifts, and these contributions are visible in the distributional comparisons (\cref{tab:ce_ape:wilcoxon}) and in the Voronoi analysis of \cref{sec:appendix:voronoi} rather than in the median APE alone.

The aggregate median also hides two failure modes that disqualify the Treatment-Mean predictor as an operational model.
The first is sensitivity to split symmetry.
Across the $50$ train/test splits used in the main text, the median absolute gap between the training-set and test-set means of $\mathit{ce\_round}$ within a treatment is $3.5\%$, but the $95$th percentile is $16\%$ and the maximum is $71\%$.
On the most asymmetric split, the Treatment-Mean median APE for the affected treatment rises from $0.050$ to $0.434$, an $8.7\times$ degradation, while the orderbook models remain within their bucket ranges.
The second is lack of transferability.
Because Treatment-Mean uses no input features, it has no mechanism to score an unseen treatment.
Under a leave-one-treatment-out protocol, in which each test treatment is predicted using the mean of the other treatments' training games, the overall median APE rises from $0.050$ to $0.078$, and for the more distinctive treatments substantially further (BBLargeCE $0.195$, FullMoreS $0.168$, FullMoreB $0.146$).
\cref{tab:cep:baseline:brittleness} summarises these figures; the per-split and per-treatment breakdowns are produced by \texttt{code/scripts/cep\_baselines\_report.py} and stored in \texttt{code/data/results/ce\_price/baseline\_diagnostics/}.

\begin{table}[htb]
\centering
\caption{\emph{Stress tests of the Treatment-Mean predictor.} The aggregate median APE over $50$ random train/test splits is close to the within-treatment dispersion of $\mathit{ce\_round}$. Two stress tests reveal that it depends on incidental symmetry of the split and fails to transfer to unseen treatments.}
\label{tab:cep:baseline:brittleness}
\small
\begin{tabular}{p{0.68\linewidth} c}
\toprule
Scenario & Median APE on CEP \\
\midrule
Treatment-Mean (aggregate, 50 splits) & 0.050 \\
Treatment-Mean on the worst-asymmetry split (sample \#41, treatment BBLargeCE; train/test mean gap +71.2\%) & 0.434 \\
Treatment-Mean under leave-one-treatment-out (novel treatment at test time) & 0.078 \\
\quad Three worst LOTO treatments: BBLargeCE = 0.195, FullMoreS = 0.168, FullMoreB = 0.146 & --- \\
\bottomrule
\end{tabular}

\end{table}

We retain the Treatment-Mean and Book-Midpoint predictors in this appendix only.
Treatment-Mean is not interpretable (it carries no coefficients or feature importances), is not transferable (it cannot score a novel treatment without retraining on that treatment's games), and is not robust to asymmetric splits.
Book-Midpoint is interpretable and transferable but is uniformly dominated.
The four models retained in the main text (EMH, CEMH, OB-RLM, GBT) combine input-responsive predictions, exposed parameters and feature importances, and acceptable robustness across splits and treatments.

\clearpage
\section{\rev{Input-Feature Ablations}}\label{sec:appendix:ablations}

We report two ablations of the inputs entering each proposed model. They test whether our conclusions rest on orderbook information or on experimental-protocol descriptors (\cref{sec:appendix:ablations:r25}), and whether the gain in performance obtained when a realised deal price is available in the current round reflects genuine predictive content or circular reuse of that price (\cref{sec:appendix:ablations:r26}). The two ablations reuse the same $50$ train/test splits as the main text and evaluate on the same test rows.

\subsection{\rev{Orderbook-only ablation}}\label{sec:appendix:ablations:r25}

\Cref{tab:model:features} lists the inputs used by each proposed model. The orderbook-only ablation removes all treatment descriptors (feedback setting and price rule), the round index and the number of observed deals from the feature vector, leaving only the running bid and ask quantiles. OB-RLM for CEP already meets this specification by design (it uses only the running quantiles). The EMH predictors (for both AE and CEP) and CEMH for AE have no orderbook inputs at all: removing their treatment/protocol inputs collapses them to a trivial constant or to the identity on the realised deal price, so the orderbook-only ablation does not apply to them. We therefore refit the remaining four of the proposed predictors in their orderbook-only variant: OB-RLM for AE (the feedback-setting grouping and the $n$ term are dropped), GBT for AE (all treatment dummies, $n$ and $\vround$ are dropped), GBT for CEP (same inputs dropped as in GBT for AE), and CEMH for CEP (the $\mathit{price\_rule}\times\mathit{feedback\_setting}$ grouping is dropped, collapsing the model to a global scalar rescaling of the realised deal price; we label this variant CEMH-global).

The refits are implemented in \texttt{code/scripts/ablation\_r25.py}. Per-split results are persisted to \texttt{code/data/results/ablations/r25/} and the aggregated comparison is reported in \cref{tab:ablation:r25}, which uses the same four (Round, Price realizations) buckets as \cref{tab:cep_ape} and \cref{tab:ae_ape}.

\begin{table}[htb]
\centering
\caption{\emph{Orderbook-only ablation.} Median APE on the same four buckets as \cref{tab:cep_ape} and \cref{tab:ae_ape}. For each model, the first row reproduces the original fit from the main text and the second row is the orderbook-only refit. GBT for CEP and OB-RLM for CEP do not change materially, while AE predictors lose accuracy after the first deal price has been observed.}
\label{tab:ablation:r25}
\small
\begin{tabular}{l l cccc}
\toprule
 & & \multicolumn{2}{c}{Round 1} & \multicolumn{2}{c}{Round $\geq 2$} \\
\cmidrule(lr){3-4}\cmidrule(lr){5-6}
Model & Variant & 0 deals & $\geq 1$ deal & 0 deals & $\geq 1$ deal \\
\midrule
\multirow{2}{*}{OB-RLM (AE)} & original & 0.216 & 0.164 & 0.120 & 0.093 \\
& orderbook-only & 0.281 & 0.164 & 0.154 & 0.136 \\
\midrule
\multirow{2}{*}{GBT (AE)} & original & 0.168 & 0.104 & 0.077 & 0.066 \\
& orderbook-only & 0.316 & 0.172 & 0.098 & 0.100 \\
\midrule
\multirow{2}{*}{CEMH (CEP)} & original & 1.000 & 0.092 & 1.000 & 0.055 \\
& orderbook-only & 1.000 & 0.092 & 1.000 & 0.054 \\
\midrule
\multirow{2}{*}{GBT (CEP)} & original & 0.135 & 0.077 & 0.099 & 0.051 \\
& orderbook-only & 0.135 & 0.077 & 0.099 & 0.051 \\
\bottomrule
\end{tabular}

\end{table}

We observe three patterns in \cref{tab:ablation:r25}. GBT for CEP is numerically identical to its main-text fit: the running bid and ask quantiles already carry the full predictive content that we attribute to the model. CEMH-global for CEP is also essentially unchanged, which reflects the fact that the per-treatment correction coefficients of CEMH are all close to one (see \cref{tab:cep:cemh:coeffs}); replacing the per-group scalar with a single global scalar therefore barely moves the predictions. The picture is different for AE. GBT for AE loses the most in the round-one no-deals bucket (from $0.168$ to $0.316$), and OB-RLM for AE degrades in three of the four buckets, most visibly in later rounds with at least one deal (from $0.093$ to $0.136$). The feedback-setting grouping and the $n$ term therefore carry real information about proximity to clearing, and the orderbook quantiles alone do not substitute for them. This justifies keeping those descriptors in the AE models of the main text, and equally confirms that on the CEP side the reported accuracy rests on the orderbook information rather than on protocol descriptors.

\subsection{\rev{Realised-price ablation}}\label{sec:appendix:ablations:r26}

According to \cref{tab:model:features}, the realised deal price only enters three of the proposed predictors. EMH for CEP is a pure identity on the last realised price and has no non-trivial variant without it. CEMH for CEP is a scalar rescaling of the realised price, so removing that input collapses the model to a per-treatment constant, which is numerically the Treatment-Mean predictor already reported in \cref{sec:appendix:cep:baselines}; we therefore reference that table here rather than duplicate the fit. The only remaining predictor is OB-RLM for AE, which we refit without the realised-price term. The refit is implemented in \texttt{code/scripts/ablation\_r26.py} and its results are stored in \texttt{code/data/results/ablations/r26/}. \Cref{tab:ablation:r26} reports the median APE in the same four buckets.

\begin{table}[htb]
\centering
\caption{\emph{Realised-price ablation for OB-RLM (AE).} Median APE on the same four buckets as \cref{tab:ae_ape}. The upper row reproduces the fit from the main text and the lower row is the refit without the realised-price input. Buckets without a realised price are unchanged by construction; buckets with a realised price lose the deal-price term.}
\label{tab:ablation:r26}
\small
\begin{tabular}{l cccc}
\toprule
 & \multicolumn{2}{c}{Round 1} & \multicolumn{2}{c}{Round $\geq 2$} \\
\cmidrule(lr){2-3}\cmidrule(lr){4-5}
Variant & 0 deals & $\geq 1$ deal & 0 deals & $\geq 1$ deal \\
\midrule
OB-RLM (AE), with realised price & 0.216 & 0.164 & 0.120 & 0.093 \\
OB-RLM (AE), without realised price & 0.215 & 0.154 & 0.127 & 0.089 \\
\bottomrule
\end{tabular}

\end{table}

The no-deals buckets are unchanged by construction because the realised-price term is zero whenever no deal has occurred in the current round. The buckets with at least one realised price are also nearly identical to the full fit: the overall median APE is $0.107$ for the full model and $0.106$ for the variant without $\vdealprice$, and the two first-round buckets remain within $0.01$ of each other. The deal-price term adds negligibly to OB-RLM AE accuracy, so the AE results in \cref{tab:ae_ape} are not driven by circular reuse of the realised price.

\section{Acronyms Table}
\begin{table}[htb!]
    \caption{List of Acronyms.}
    \label{tab:acronyms}
    \centering
    \begin{tabular}{ll}
        \toprule
        \textbf{Acronym} & \textbf{Description} \\
        \midrule
        AE & Allocative Efficiency \\
        APE & Absolute Percentage Error \\
        BB & Black Box (feedback setting) \\
        CE & Competitive Equilibrium \\
        CEMH & Corrected Efficient Market Hypothesis \\
        CEP & Competitive Equilibrium Price \\
        CV & Cross-Validation \\
        EMH & Efficient Market Hypothesis \\
        FS & Feedback Setting \\
        GBT & Gradient Boosted Trees \\
        GOT & Gains of Trade \\
        IQR & Interquartile Range \\
        LOB & Limit Order Book \\
        LOTO & Leave-One-Treatment-Out (cross-validation) \\
        MAPE & Median Absolute Percentage Error \\
        ML & Machine Learning \\
        MMK & Matchmaker Keeps (price rule) \\
        OB-RLM & Orderbook Robust Linear Model \\
        PDP & Partial Dependence Plot \\
        PR & Price Rule \\
        RLM & Robust Linear Model \\
        RMSE & Root Mean Squared Error \\
        SHAP & SHapley Additive exPlanations \\
        ZI & Zero-Intelligence (agents) \\
        \bottomrule
    \end{tabular}
\end{table}

\end{document}